\documentclass[journal]{IEEEtran} 
\usepackage{amsmath,amsfonts}
\usepackage{algorithmic}
\usepackage{algorithm}
\usepackage{array}
\usepackage[caption=false,font=normalsize,labelfont=sf,textfont=sf]{subfig}
\usepackage{textcomp}
\usepackage{float}
\usepackage{stfloats}
\usepackage{url}
\usepackage{verbatim}
\usepackage{makecell}
\usepackage{graphicx}
\usepackage{cite}
\usepackage{booktabs}
\usepackage{multirow}
\usepackage{pifont}
\usepackage{xcolor}
\usepackage{bm}
\usepackage{booktabs}   % For professional-quality rules (\toprule, \midrule, \bottomrule)
\usepackage{tabularx}   % For the tabularx environment (tables with fixed total width)
\usepackage{ragged2e}   
\usepackage{threeparttable}% Provides\Centering, more robust than \centering

\usepackage{xcolor}
\definecolor{bestblue}{RGB}{0, 70, 140}    
\definecolor{secondblue}{RGB}{100, 140, 200}

\newcolumntype{C}{>{\Centering\arraybackslash}X} 
\newcolumntype{L}{>{\RaggedRight\arraybackslash}X} 

\usepackage[colorlinks=true,linkcolor=black,citecolor=black,urlcolor=magenta]{hyperref}

\begin{document}
\bstctlcite{IEEEexample:BSTcontrol}
\title{
Spline Policy: A Structured Representation for Robot Policies
% Spline Policy: \\
% \scalebox{0.8}{Revisiting and Reinterpreting Splines for Manipulation Planning}
}

% \author{Author Names Omitted for Anonymous Review}

\author{
Mengze Tian$^{1,\dagger}$,
Yiming Li$^{2,1,\dagger}$,
Sichao Liu$^{1}$,
Auke Ijspeert$^{1}$,
Sylvain Calinon$^{2,1}$%
\thanks{$^{1}$École Polytechnique Fédérale de Lausanne (EPFL), Lausanne, Switzerland.}
\thanks{$^{2}$Idiap Research Institute, Martigny, Switzerland.}
\thanks{$^{\dagger}$These authors contributed equally to this work.}
\thanks{Corresponding author: Yiming Li (yiming.li@epfl.ch).}
}


\maketitle

\begin{abstract}
Modern imitation-learning policies for robot manipulation often represent actions as fixed-resolution action chunks, which are simple and effective but expose limited geometric and temporal structure before execution. This paper studies Spline Policy (SP), a structured representation that replaces action chunks with spline parameters while keeping the policy backbone unchanged. The predicted spline can be decoded as a compact continuous trajectory, queried at different temporal resolutions, constrained or edited in parameter space, and passed to downstream controllers. For quadratic spline outputs, the same representation can also be converted into a state-dependent vector field through an analytical distance-field construction. Under the regularity and projection assumptions of this construction, the induced dynamics do not increase the distance to the generated spline, yielding a principled local corrective mechanism around the predicted motion. The spline output further supports uncertainty propagation from observations to spline parameters, trajectories, and flow fields, and can be combined with classical control mechanisms such as null-space collision avoidance without retraining the policy backbone. We instantiate SP with diffusion, flow-matching, transformer-based, and vision-language-action backbones. Experiments in low-dimensional motion learning, simulated manipulation under matched backbones, dexterous manipulation, and real-robot case studies show that SP remains compatible with modern policy learners while exposing useful motion-structure properties, including compact decoding, temporal resampling, local correction around predicted motions, uncertainty evaluation, and controller compatibility.
Video: \url{https://youtu.be/ct_44GcgX-I}.
Code: \url{https://github.com/mengze3/spline_policy}.
\end{abstract}


\section{Introduction}

Imitation learning has become a central paradigm for robot manipulation, enabling policies to be acquired directly from demonstrations rather than being engineered task by task. Recent data-driven policy models have made imitation learning effective in settings involving multimodal behaviors and high-dimensional observations, including Action Chunking Transformer (ACT), Diffusion Policy (DP), Flow Matching Policy (FMP), Vision-Language-Action (VLA) models, and Vision-Action (VA) models~\cite{zhao2023learning,reuss2023goal,chi2024diffusionijrr,lipman2023flow, braun2024riemannian,zhang2024affordance, physicalintelligence2025pi05,zitkovich2023rt2,pai2025mimic,han2025hierarchically, li2026causal,kim2026cosmos}. 
Many of these approaches represent policies through \emph{action chunks}: fixed-horizon sequences of actions predicted by high-capacity neural models. 
This interface is simple, but the predicted object is usually a fixed-resolution sequence of discrete points. 
As a result, geometric and temporal structures that are useful for execution, such as continuity, derivative information, boundary conditions, and temporal resampling, are not explicitly exposed before control.

In parallel, the learning-from-demonstration literature has long emphasized the value of \emph{movement primitives} as structured motion representations~\cite{ijspeert2013dmp,ijspeert2002movement,billard2008robot,paraschos2013promp,calinon2019mixture,otto2023mp3,scheikl2024movement,zhou2025beast,yang2026abpolicyasynchronousbsplineflow,carvalho2025motion}. 
Compared with fixed-resolution action chunks, movement primitives provide compact parameterizations, support smooth, temporally consistent trajectory generation, and offer natural compatibility with control, optimization, and trajectory editing. 
Modern policy learning and movement primitives therefore provide complementary strengths: the former offers scalable perception-conditioned and multimodal behavior modeling, while the latter exposes geometric, temporal, and control-relevant motion structure. 
This motivates studying the policy-output interface as a point of connection between these two perspectives.

In this paper, we study splines as a structured output interface for modern robot policies and refer to the resulting formulation as Spline Policy (SP). 
Splines are a natural candidate for this role because they encode continuous trajectories using a small set of geometric parameters, such as control points, while remaining simple and model-agnostic. 
Their local support facilitates local trajectory editing, and continuity, boundary conditions, derivative constraints, and temporal resampling can often be handled directly in parameter space~\cite{calinon2019mixture,saveriano2023dynamic}. 
These properties are well established in classical trajectory representation. 
Our focus is to examine how they can be used as a structured representation alongside fixed-resolution action chunks in modern robot policy learning.

Given an observation, the policy backbone predicts spline parameters rather than directly outputting a discrete action sequence. 
The predicted spline can then be decoded as a continuous trajectory, queried at different control frequencies, constrained or edited in parameter space, and passed to downstream controllers. 
In this sense, SP provides a lightweight bridge between expressive neural policy learning and structured motion representations: the backbone remains responsible for perception, multimodality, and task-level prediction, while the spline output exposes a compact continuous motion object before execution. 
This object is compatible with temporal resampling, local flow construction, uncertainty visualization, and controller integration.

Beyond direct trajectory decoding, SP can also be realized as a flow-field policy by constructing a state-dependent dynamical system from the predicted spline. 
This builds on recent work showing that quadratic splines admit an analytical distance-field construction, which can be used to define a dynamical system with convergence properties under appropriate regularity and projection assumptions~\cite{li2025movement}. 
In this realization, the neural policy does not learn an unconstrained vector field directly. 
Instead, it predicts the geometric support of the motion through the same imitation learning method, while the spline-to-field transformation supplies the closed-loop vector field structure. 
During execution, the robot follows the velocity induced at its current state rather than simply replaying a time-indexed trajectory. 
This provides a structured mechanism for correcting deviations caused by physical perturbations, tracking errors, or other execution-time disturbances around the generated motion.

The corrective property of this flow-field realization is local to the predicted spline. 
Under the stated regularity and projection assumptions, the induced dynamics do not increase the distance to the generated spline, and the terminal point becomes an equilibrium when a zero terminal tangent is imposed. 
Thus, the flow-field realization provides a closed-loop execution mechanism around the predicted spline, while task success still depends on the quality of the spline predicted by the learned policy. 
The same spline output also supports uncertainty propagation from observations to spline parameters and flow fields, and can be integrated with classical controllers such as null-space collision avoidance without retraining the policy backbone.

\begin{figure*}[t]
    \centering
    \includegraphics[width=0.95\textwidth]{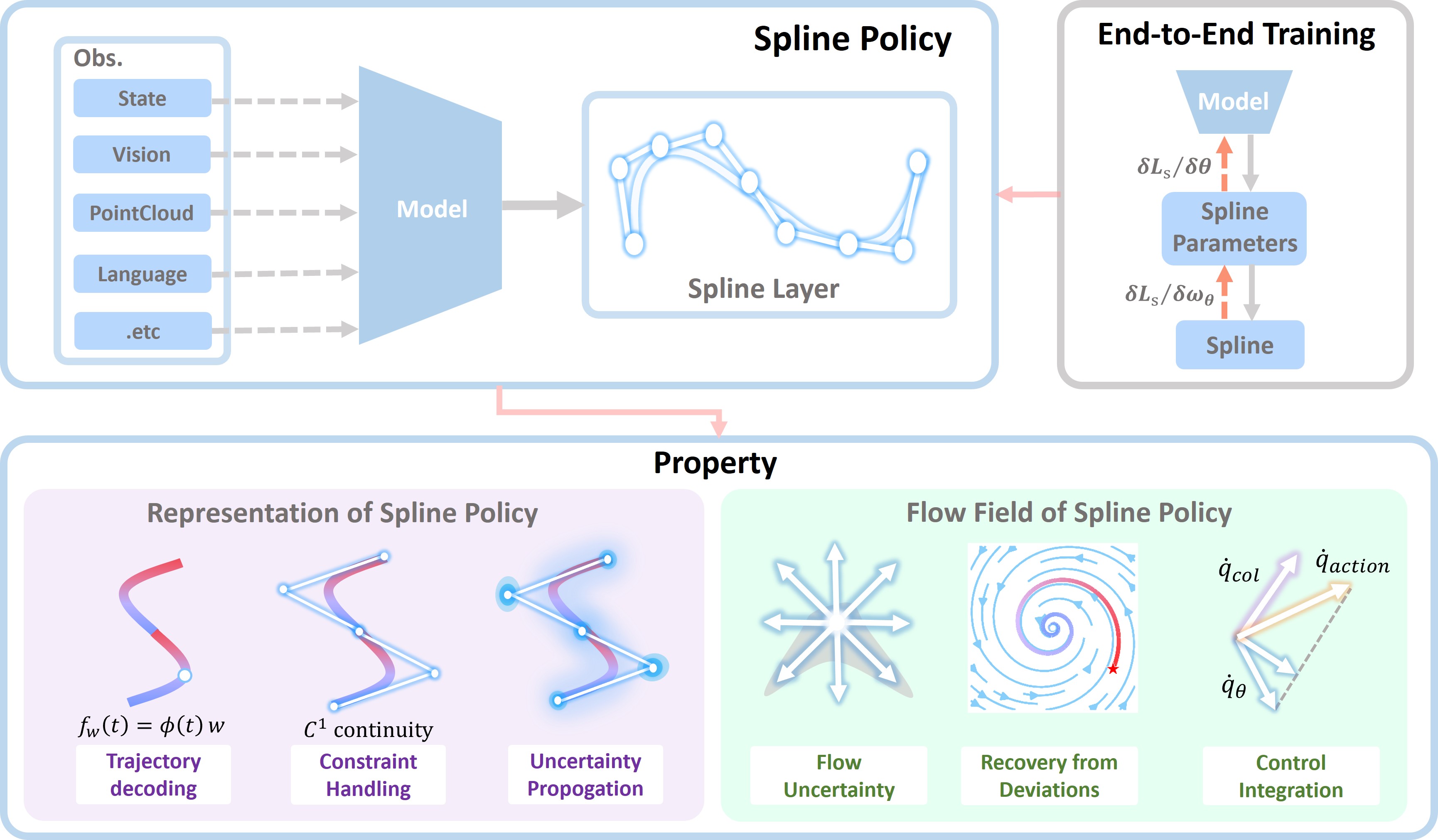}
    \caption{
Overview of SP as a structured output interface.
\textbf{Top-left:} An observation is processed by an existing policy backbone, while the output representation is replaced by spline parameters.
\textbf{Top-right:} Directly predicting spline parameters via end-to-end training.
\textbf{Bottom:} The same spline output supports complementary uses. As a trajectory representation, it provides compact, smooth temporal decoding; as a flow-field realization, it can be transformed into a local state-dependent correction mechanism via a spline-to-field construction.
}
    \label{fig:method}
    \vspace{-15pt}
\end{figure*}

Building on these observations, we instantiate SP with several modern backbones, including ACT, DP, FMP, and VLA models. 
In each case, the policy backbone is preserved, while the output representation, prediction target, and corresponding loss are changed from action chunks to spline parameters. 
We use simulated manipulation benchmarks as the primary evaluation of SP as a structured layer under matched backbones. 
The low-dimensional and real-world studies then illustrate mechanism- and deployment-level affordances of the same spline output, rather than serving as separate claims of policy-learning superiority.

In summary, the main contributions are:
\begin{itemize}
    \item We exploit concatenated splines as a structured representation for modern imitation learning policies, enabling fixed-resolution action chunks to be represented by compact spline parameters with smooth, temporally flexible trajectory decoding.

    \item We introduce a flow-field realization of Spline Policy, where the predicted spline induces a state-dependent vector field for closed-loop execution, local perturbation recovery, uncertainty propagation, and control integration.

    \item We instantiate and evaluate SP across multiple policy backbones, input modalities, and robot manipulation settings, including simulated manipulation benchmarks, dexterous manipulation tasks, and real-world robotic experiments.
\end{itemize}

\section{Related Work}

\subsection{Movement primitives} 
Movement primitives (MPs)~\cite{calinon2019mixture} provide compact and parameter-efficient representations of robotic trajectories, enabling smooth interpolation and compatibility with classical control pipelines. 
Representative approaches include Dynamical Movement Primitives (DMPs)~\cite{ijspeert2002movement,saveriano2023dynamic}, Task-Parameterized Gaussian Mixture models \cite{calinon2016tp}, Probabilistic Movement Primitives (ProMPs)~\cite{paraschos2013promp}, which encode demonstrations into reusable trajectory modules, and piecewise-polynomial representations for motion planning~\cite{mellinger2011minimum,richter2016polynomial,gao2018bspline}.
Recent studies within learning frameworks~\cite{otto2023mp3,scheikl2024movement,zhou2025beast,yang2026abpolicyasynchronousbsplineflow} extend this paradigm by replacing discrete trajectory representations with MPs, leveraging their structured formulation for policy learning.

In contrast to prior works that incorporate movement primitives within specific model classes or use them mainly as auxiliary trajectory encodings, we focus on their role as a policy-output interface. 
SP is designed to be inserted into modern policy-learning pipelines without changing the backbone architecture, allowing expressive perception-conditioned models to output a structured motion representation.

\vspace{-10pt}
\subsection{Dynamical systems}
Dynamical system representations describe robot motion through differential equations, often in the form of a vector field over the state or configuration space~\cite{khansari2011learning,khansari2012dynamical,khansari2014learning,ratliff2018riemannian,urain2020imitationflow,bahl2020neural,beik2024neural,jiang2025streaming,li2025elastic,simmoteit2025diffeomorphic}. 
In learning from demonstration, many approaches use autonomous vector fields, where the current state determines the motion direction without requiring an explicit time index. 
Such formulations can reduce dependence on time synchronization, support reactive responses to perturbations, and provide continuous encodings of motion.
Examples include SEDS~\cite{khansari2011learning}, CLF-DM~\cite{khansari2014learning}, Imitation Flow~\cite{urain2020imitationflow}, and Neural Contractive Dynamical Systems (NCDS)~\cite{beik2024neural}, which ensure stability or adaptability by learning or imposing constraints on the system dynamics.

These approaches most often rely on explicit dynamical-system formulations that are not directly aligned with modern policy-learning pipelines, which commonly learn observation-conditioned action trajectories. 
Connecting such formulations to high-dimensional observations, multimodal behaviors, or action-chunk policy backbones often requires additional modeling or integration choices.
Recent work~\cite{li2025movement,li2026geometryaware} shows that spline geometry can be used to construct distance fields analytically, whose gradients can be exploited to construct dynamical systems. 
Motivated by this connection, we use expressive policy learning to predict spline geometry, from which a flow field can be obtained through a spline-to-field transformation. 
This separates the learning problem from the structure-inducing transformation: the neural policy predicts the geometric support of the motion, while the vector-field structure is supplied by the spline-to-field construction.

\subsection{Imitation learning} 
In manipulation, imitation learning (IL)~\cite{calinon2007learning,billard2008robot,osa2018algorithmic,calinon2019mixture} has played a central role due to its ability to learn policies directly from expert demonstrations, bypassing the need for task-specific engineering.
Many approaches in IL rely on movement primitives~\cite{ijspeert2002movement,ijspeert2002movement,paraschos2013promp} and dynamical systems~\cite{khansari2011learning,khansari2012dynamical,khansari2014learning}, demonstrating their effectiveness in capturing geometric and dynamical motion properties.
More recent policy learning frameworks now include ACT, DP, FMP, VLA, and other VA models~\cite{reuss2023goal,chi2024diffusionijrr,lipman2023flow,braun2024riemannian,zhang2024affordance,zhao2023learning,physicalintelligence2025pi05,zitkovich2023rt2,pai2025mimic,han2025hierarchically,li2026causal,kim2026cosmos}.
These approaches leverage powerful function approximators to model complex, high-dimensional, and multimodal behavior distributions directly from demonstration data, thereby improving the expressiveness and scalability of imitation learning.

SP is complementary to these policy-learning backbones. 
Rather than introducing a new imitation learning architecture, the proposed approach is to modify on-the-fly the representation of the predicted action object. 
This allows modern policies to retain their modeling capacity while exposing motion structure that is useful for decoding, replanning, uncertainty propagation, and control.

\section{Spline Policy as a Structured Representation}

\subsection{Problem Formulation}

Let $\mathbf{o}$ denote an observation, which may include robot states, visual inputs, point clouds, or language instructions. 
In many modern robot policies, the output is parameterized as an action chunk, where a model $\epsilon_\theta$ predicts a finite sequence of actions
\begin{equation}
\mathbf{a}_{1:N} = \epsilon_\theta(\mathbf{o}),
\label{eq:action_chunk}
\end{equation}
for a fixed horizon $N$. 
The model $\epsilon_\theta$ may be a diffusion model, a flow-matching model, a transformer policy, or a VLA backbone.

Instead of changing the policy backbone, SP changes the representation of the predicted action object. 
Given an observation $\mathbf{o}$, the same model predicts a set of concatenated spline parameters
\begin{equation}
\mathbf{w}_\theta(\mathbf{o}) = \epsilon_\theta(\mathbf{o}).
\label{eq:spline_param}
\end{equation}
The corresponding continuous trajectory is obtained by evaluating the concatenated spline basis
\begin{equation}
\mathbf{f}_{\mathbf{w}_\theta(\mathbf{o})}(t)
=
\boldsymbol{\phi}(t)\,\mathbf{w}_\theta(\mathbf{o}),
\label{eq:spline_decode}
\end{equation}
where $\boldsymbol{\phi}(t)$ denotes the spline basis and 
$\mathbf{f}_{\mathbf{w}_\theta(\mathbf{o})}(t)$ is the decoded trajectory at time $t$~\cite{calinon2019mixture}. 
Under this formulation, the policy outputs spline parameters rather than a fixed-resolution sequence of discrete actions. 
The decoded trajectory is continuous and differentiable, with a number of predicted variables smaller than the typical number of actions considered in dense action chunks. 
This provides a compact representation while keeping the policy-learning problem compatible with existing backbones, as illustrated in Fig.~\ref{fig:method}.

In practice, the trajectory can be represented by $K$ spline segments, each defined over a normalized local time interval and parameterized by a small number of control variables. 
Continuity constraints of order $C^0$, $C^1$, or $C^2$ can be imposed across neighboring segments depending on the desired level of smoothness. 
The trajectory-decoding view is not tied to a specific spline family: different spline bases or orders can be used depending on the task, controller, and desired smoothness. 
In our implementation, we mainly use piecewise quadratic splines, which provide additional analytic properties, while the flow-field realization is introduced separately in Sec.~\ref{sec:flow_field}.

\subsection{Structured Decoding and Operations}
\label{subsec:structured_operations}

Once the policy predicts spline parameters, the resulting object exposes several standard operations that are difficult to access directly from an unconstrained fixed-resolution action chunk. 
We treat these operations as interface affordances rather than as separately validated algorithms. 
We describe three such operations: temporally flexible decoding, constraint handling, and uncertainty propagation.

To make these operations concrete, we use the piecewise quadratic Bernstein representation adopted in our implementation. 
For segment $i$, with local phase $\tau \in [0,1]$, the decoded trajectory is
\begin{equation}
\mathbf{f}_{\theta,i}(\tau)
=
(1-\tau)^2 \mathbf{w}_i^1
+
2(1-\tau)\tau \mathbf{w}_i^2
+
\tau^2 \mathbf{w}_i^3,
\quad \tau \in [0,1],
\label{eq:quadratic_bernstein_segment}
\end{equation}
where $\mathbf{w}_i^1,\mathbf{w}_i^2,\mathbf{w}_i^3$ are the control points of the segment. 
The same discussion applies to other spline bases with the corresponding basis functions and control variables.

\textbf{Trajectory decoding:}
Splines make the predicted action object compact and continuously decodable, as illustrated in Fig.~\ref{fig:method}. 
A single policy prediction defines a continuous curve that can be evaluated at different temporal resolutions during execution, without retraining the policy or changing the output dimension. 
This decouples the learned prediction from the controller query rate and allows the same spline parameters to be used for coarse planning, high-rate control, or visualization. 
Since the spline is differentiable, velocity and acceleration can also be obtained analytically when required by downstream controllers. 
When continuity constraints are imposed between segments, the structure can easily impose a desired level of continuity ($
C^0, C^1, \ldots$) when action chunks are repeatedly replanned.

\textbf{Constraint handling:}
Because SP exposes the predicted motion through spline parameters, common geometric and temporal constraints can be handled at the level of control variables. 
In this work, we mainly use boundary and continuity constraints, while the same representation also provides a convenient interface for derivative bounds and convex workspace or joint-space constraints.

For the quadratic segment in Eq.~\ref{eq:quadratic_bernstein_segment}, boundary constraints can be imposed by fixing endpoint control points, such as
$\mathbf{w}_1^1=\mathbf{x}_{\mathrm{start}}$ or
$\mathbf{w}_K^3=\mathbf{x}_{\mathrm{goal}}$. 
Continuity constraints between neighboring segments become linear equality constraints on boundary control points. 
For example, $C^0$ continuity requires
\begin{equation}
\mathbf{w}_i^3 = \mathbf{w}_{i+1}^1,
\end{equation}
while $C^1$ continuity, for segment durations $\Delta t_i$ and $\Delta t_{i+1}$, can be written as
\begin{equation}
\frac{\mathbf{w}_i^3-\mathbf{w}_i^2}{\Delta t_i}
=
\frac{\mathbf{w}_{i+1}^2-\mathbf{w}_{i+1}^1}{\Delta t_{i+1}} .
\end{equation}
The zero-terminal-tangent condition used in the flow-field realization is another boundary constraint and can be imposed by setting
\begin{equation}
\mathbf{w}_K^2 = \mathbf{w}_K^3 .
\end{equation}

The spline parameterization also gives a simple way to express sufficient conditions for inequality constraints. 
For Bernstein and B-spline representations, the basis functions are nonnegative and form a partition of unity, so each segment lies in the convex hull of its control points. 
Consequently, if the control points of a segment are constrained to lie in a convex set $\mathcal{C}_i$, then the entire segment remains inside that set:
\begin{equation}
\mathbf{w}_i^j \in \mathcal{C}_i,\quad j=1,2,3
\quad \Rightarrow \quad
\mathbf{f}_{\theta,i}(\tau) \in \mathcal{C}_i,\ \forall \tau \in [0,1].
\end{equation}
This property provides a sufficient condition for box constraints, joint limits, or convex safe corridors. 
Derivative constraints can similarly be expressed through control-point differences. 
For the quadratic segment in Eq.~\ref{eq:quadratic_bernstein_segment}, its derivative with respect to the local phase $\tau$ is simply
\begin{equation}
\dot{\mathbf{f}}_{\theta,i}(\tau)
=
2(1-\tau)(\mathbf{w}_i^2-\mathbf{w}_i^1)
+
2\tau(\mathbf{w}_i^3-\mathbf{w}_i^2),
\end{equation}
where the dot denotes differentiation with respect to $\tau$. 
Thus, derivative bounds can be imposed through constraints on the corresponding control-point differences. 

\begin{figure}[htbp]
    \centering
    \includegraphics[width=0.95\linewidth]{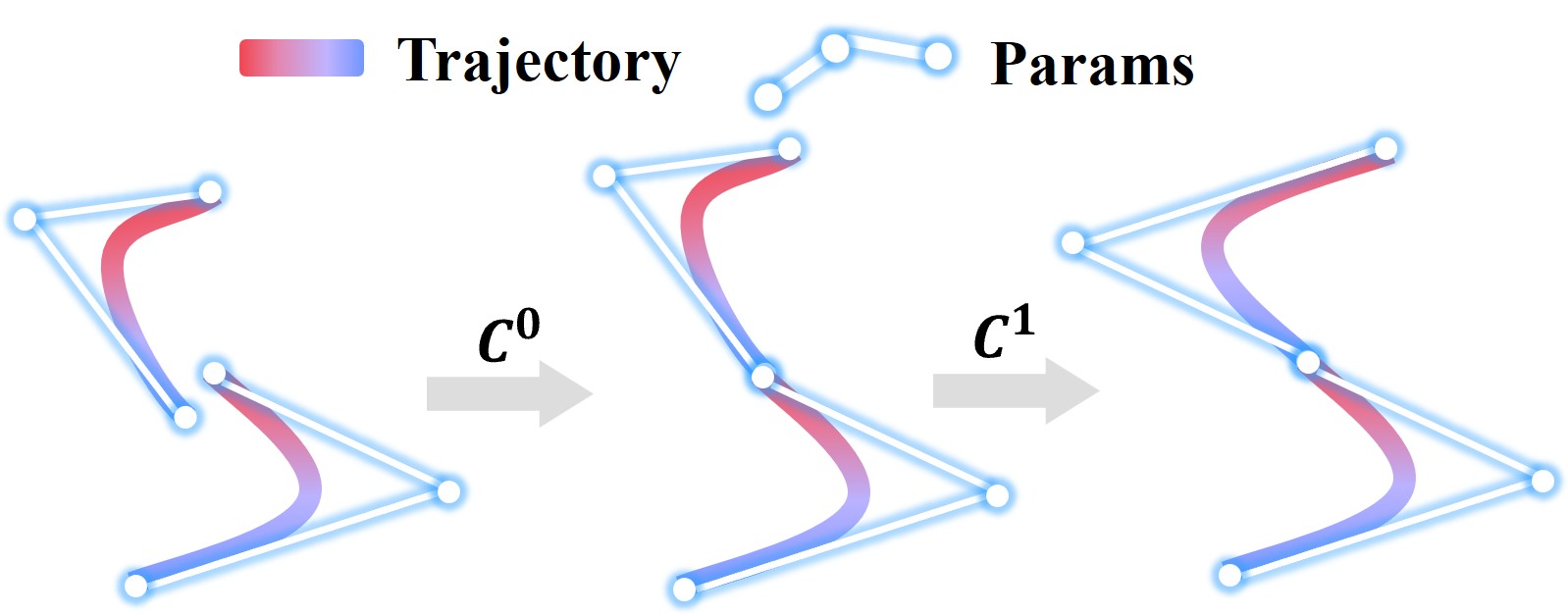}
    \caption{
    Boundary constraint. 
    \textbf{Left:} No continuity constraints. 
    \textbf{Middle:} Enforcing $C^0$ positional continuity connects neighboring segments at the junction. 
    \textbf{Right:} Enforcing $C^1$ continuity additionally ensures velocity consistency, resulting in smoother transitions.
    }
    \label{fig:mechanism_continuity}
    \vspace{-10pt}
\end{figure}

\begin{figure}[htbp]
    \centering
    \includegraphics[width=0.95\linewidth]{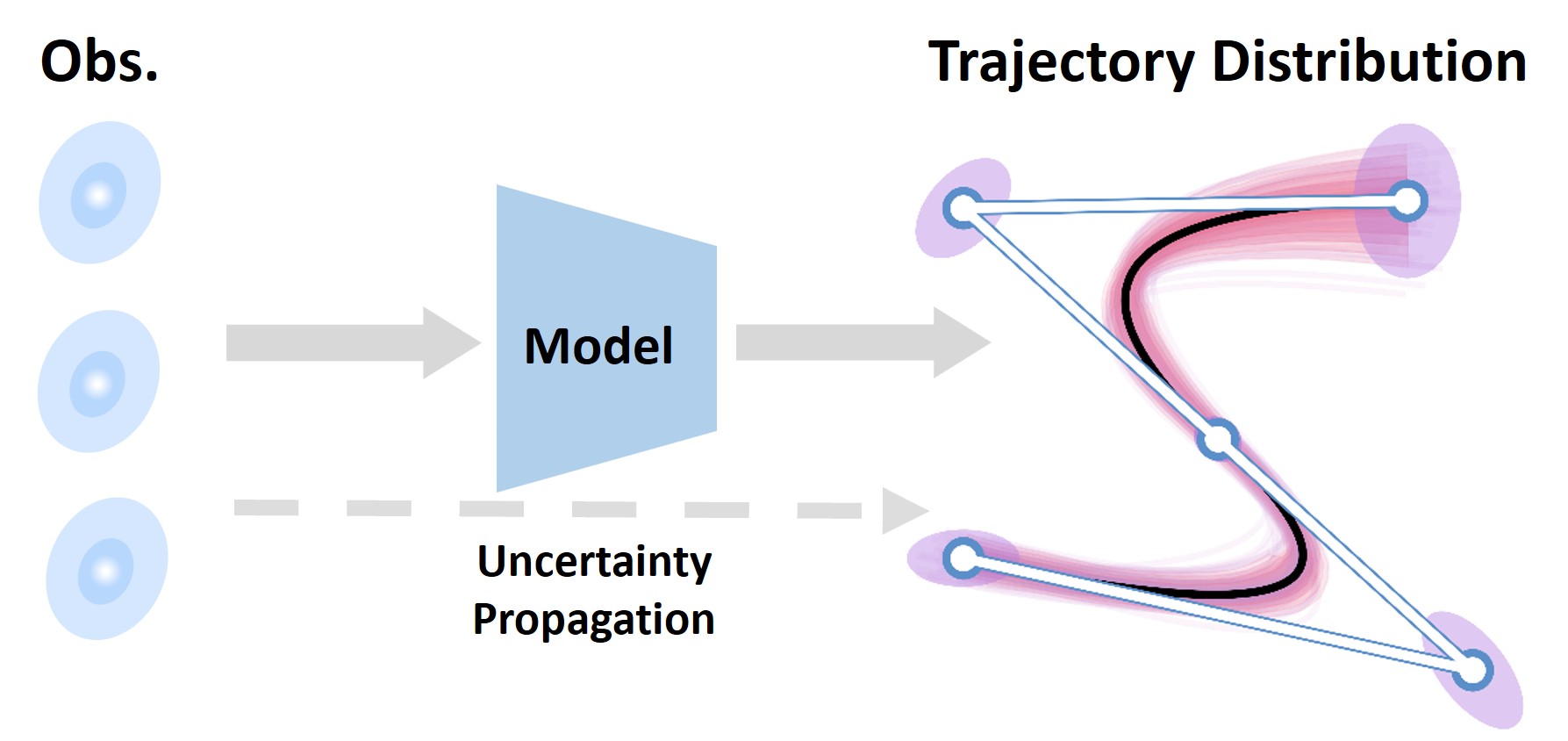}
    \caption{
    Uncertainty propagation.
    \textbf{Left:} Noisy observations.
    \textbf{Middle:} The model infers spline parameters.
    \textbf{Right:} The input uncertainty is propagated, yielding a distribution over trajectories.
    The black curve indicates the mean trajectory, and the surrounding clusters illustrate the distribution of trajectories.
    }
    \label{fig:spline_uncertainty}
    \vspace{-5pt}
\end{figure}

\textbf{Uncertainty propagation:}
\label{sec:uncertainty_spline}
In manipulation tasks, observation noise or perception ambiguity can lead to different plausible motion predictions. 
Because SP represents the predicted motion through a compact set of spline parameters, this uncertainty can be represented at the parameter level and then propagated analytically to the decoded trajectory. 
This provides a computationally light (just linear transformations with matrix multiplications) to evaluate how uncertainty in the input affects the predicted motion, without changing the policy backbone or the spline decoder, as illustrated in Fig.~\ref{fig:spline_uncertainty}.

Assume that the observation $\mathbf{o}$ follows an underlying distribution that captures sensing noise or perception uncertainty. 
Since the spline parameters are obtained through the mapping $\mathbf{w}_\theta(\mathbf{o})$, they inherit the uncertainty in $\mathbf{o}$. 
We compute the resulting distribution over spline parameters as a Gaussian
\begin{equation}
    \mathbf{w}_\theta(\mathbf{o}) \sim \mathcal{N}(\boldsymbol{\mu}_{w}, \boldsymbol{\Sigma}_{w}),
\label{eq:spline_param_uncertainty}
\end{equation}
where $\boldsymbol{\mu}_{w}$ and $\boldsymbol{\Sigma}_{w}$ denote the mean and covariance of the parameter distribution. 
This Gaussian model is used as a tractable approximation; when the underlying behavior distribution is strongly multimodal, mixture or sampling-based representations can be used instead.

For a given time $t$, the trajectory distribution is obtained by propagating this uncertainty through the spline basis
\begin{equation}
    \mathbf{f}_{\mathbf{w}_\theta(\mathbf{o})}(t)
    \sim
    \mathcal{N}\Bigl(
        \boldsymbol{\phi}(t)\boldsymbol{\mu}_{w},\;
        \boldsymbol{\phi}(t)\boldsymbol{\Sigma}_{w}\boldsymbol{\phi}(t)^\top
    \Bigr).
\label{eq:spline_traj_uncertainty}
\end{equation}
Equivalently, for a set of queried time instants, the same propagation can be applied using the corresponding stacked spline basis matrices. The resulting trajectory distribution provides a diagnostic signal for the predicted motion by highlighting time intervals or spatial regions where the policy output is sensitive to observation uncertainty. 
This can help downstream execution behave more conservatively when the predicted motion is uncertain, for example, by slowing down, replanning, or collecting additional observations.
% In practice, we estimate $\boldsymbol{\mu}_{w}$ and $\boldsymbol{\Sigma}_{w}$ via Monte Carlo sampling by drawing observations from their distribution and computing the corresponding samples of $\mathbf{w}_\theta(\mathbf{o})$. 
% This procedure is compatible with batch computation and can be efficiently parallelized.

\subsection{Policy Integration}
\label{subsec:policy_integration}

SP is integrated into modern policy-learning pipelines by training the policy to output spline parameters end to end. 
The perception encoder and policy backbone are kept unchanged, while the output head, prediction target, and decoding layer are adapted to the spline representation. 
Thus, the main modification is localized to the policy-output interface rather than to the architecture used for perception or sequence modeling.

For supervised policy backbones, training can be performed by aligning the decoded spline trajectory with the demonstrated trajectory $\mathbf{f}_d(t)$ over sampled time instants $\{t_i\}_{i=1}^{N}$:
\begin{equation}
\mathcal{L}_{s}
=
\frac{1}{N}
\sum_{i=1}^{N}
\left\|
\mathbf{f}_{\mathbf{w}_\theta(\mathbf{o})}(t_i)
-
\mathbf{f}_d(t_i)
\right\|^2 .
\label{eq:spline_loss}
\end{equation}
Gradients can be back-propagated through the spline representation to the policy parameters:
\begin{equation}
\frac{\partial \mathcal{L}_{s}}{\partial \theta}
=
\frac{\partial \mathcal{L}_{s}}{\partial
\mathbf{f}_{\mathbf{w}_\theta(\mathbf{o})}}
\;
\frac{\partial
\mathbf{f}_{\mathbf{w}_\theta(\mathbf{o})}
}{\partial \mathbf{w}_\theta(\mathbf{o})}
\;
\frac{\partial \mathbf{w}_\theta(\mathbf{o})}{\partial \theta}.
\label{eq:chain_rule}
\end{equation}
This allows trajectory-level losses or differentiable constraint penalties to be composed with the policy-learning objective when desired.

SP is thus model-agnostic: it can be attached to different policy backbones by preserving the original architecture and replacing the action chunk output head and target with spline-parameter predictions. 
This applies to diffusion, flow-matching, transformer, VLA-style policies, and other action prediction models. 
The training objective is otherwise unchanged, and the decoded spline is used for execution.
A trained policy can also be converted post hoc by fitting a spline to its predicted action sequence. 
This yields a continuous trajectory, but only reparameterizes the output after prediction; end-to-end spline prediction instead exposes the structured output space during training and optimizes through the spline decoder.

\section{Flow-Field Realization of Spline Policy}
\label{sec:flow_field}

Beyond direct trajectory decoding, the spline predicted by SP can also be used to construct a spatial flow field for closed-loop execution.
We denote here the decoded spline in \eqref{eq:spline_decode} by $\mathbf{f}_\theta(t)$ for brevity.
Instead of tracking $\mathbf{f}_\theta(t)$ only as a time-parameterized trajectory, we regard it as a geometric object embedded in the state space and construct a distance field around it.
The distance field and its associated projection map are then used to define a state-dependent flow field.
This provides a closed-loop execution mode derived from the same spline output. 
In this work, we instantiate this principle with concatenated quadratic splines (piecewise continuous curves), for which the projection map and distance-field gradient admit an analytical construction~\cite{li2025movement}.
For higher-order splines or other spline families, this construction remains applicable, but the flow-field realization would require numerical projection or optimization-based closest-point computation. 

\subsection{From Spline to Flow Field}

Let $\mathbf{f}_\theta$ denote the spline predicted by SP, and let 
$\mathcal{P}_{\mathbf{f}_\theta}$ be the closest-point projection from a query state 
$\mathbf{x}\in\mathcal{X}$ to the spline. 
The projection returns the corresponding phase and projected point:
\begin{equation}
    \bigl(t_\theta(\mathbf{x}), \mathbf{x}_{\mathrm{proj}}\bigr)
    =
    \mathcal{P}_{\mathbf{f}_\theta}(\mathbf{x}),
    \qquad
    \mathbf{x}_{\mathrm{proj}}
    =
    \mathbf{f}_\theta(t_\theta(\mathbf{x})).
    \label{eq:spline_projection_operator}
\end{equation}
At regular points where the projection is unique and differentiable, it induces the distance
$d_\theta(\mathbf{x})=\|\mathbf{x}-\mathbf{x}_{\mathrm{proj}}\|$ and the normal direction
\begin{equation}
    \mathbf{n}_\theta(\mathbf{x})
    =
    \frac{\mathbf{x}-\mathbf{x}_{\mathrm{proj}}}
    {\|\mathbf{x}-\mathbf{x}_{\mathrm{proj}}\|}.
\end{equation}
For quadratic splines, these quantities can be computed analytically, as detailed in App.~\ref{app:spline_to_distance_field}. 
For other spline families, they can be obtained numerically by solving a one-dimensional closest-point problem along the spline.

The flow field is constructed by combining attraction toward the spline and progression along the spline
\begin{equation}
    \mathbf{F}_\theta(\mathbf{x})
    =
    \mathbf{v}_{\mathrm{att}}(\mathbf{x})
    +
    \mathbf{v}_{\mathrm{prog}}(\mathbf{x}),
    \label{eq:sp_flow_decomposition}
\end{equation}
with
\begin{equation}
\begin{aligned}
    \mathbf{v}_{\mathrm{att}}(\mathbf{x})
    &=
    \alpha(\mathbf{x}) \;
    \mathbf{n}_\theta(\mathbf{x}),
    \\
    \mathbf{v}_{\mathrm{prog}}(\mathbf{x})
    &=
    \beta(\mathbf{x}) \;\dot{\mathbf{f}}_\theta(\mathbf{x}).
    \label{eq:attraction_progression}
\end{aligned}
\end{equation}
Here, $\alpha(\cdot)\le 0$ and $\beta(\cdot)\ge 0$ control the relative strength of attraction and tangential progression. 
The weight function $\alpha(\mathbf{x})$ is set so that it reaches zero when $d_\theta(\mathbf{x})=0$ (typically, to emulate a spring behavior whose strength is proportional to the distance to the curve). 
The weight function $\beta(\mathbf{x})$ is set so that it reaches zero when $t_\theta(\mathbf{x})=T$ (stopping the motion at the end of the curve).

With \eqref{eq:sp_flow_decomposition}, states away from the spline are driven back toward the predicted geometric path, while states near the curve mainly follow the tangential direction. 
The generated policy, therefore, predicts the curve geometry rather than an unconstrained vector field through a distance function.

\subsection{Local Correction and Perturbation Recovery}

\begin{figure}[!h]
    \centering
    \includegraphics[width=\linewidth]{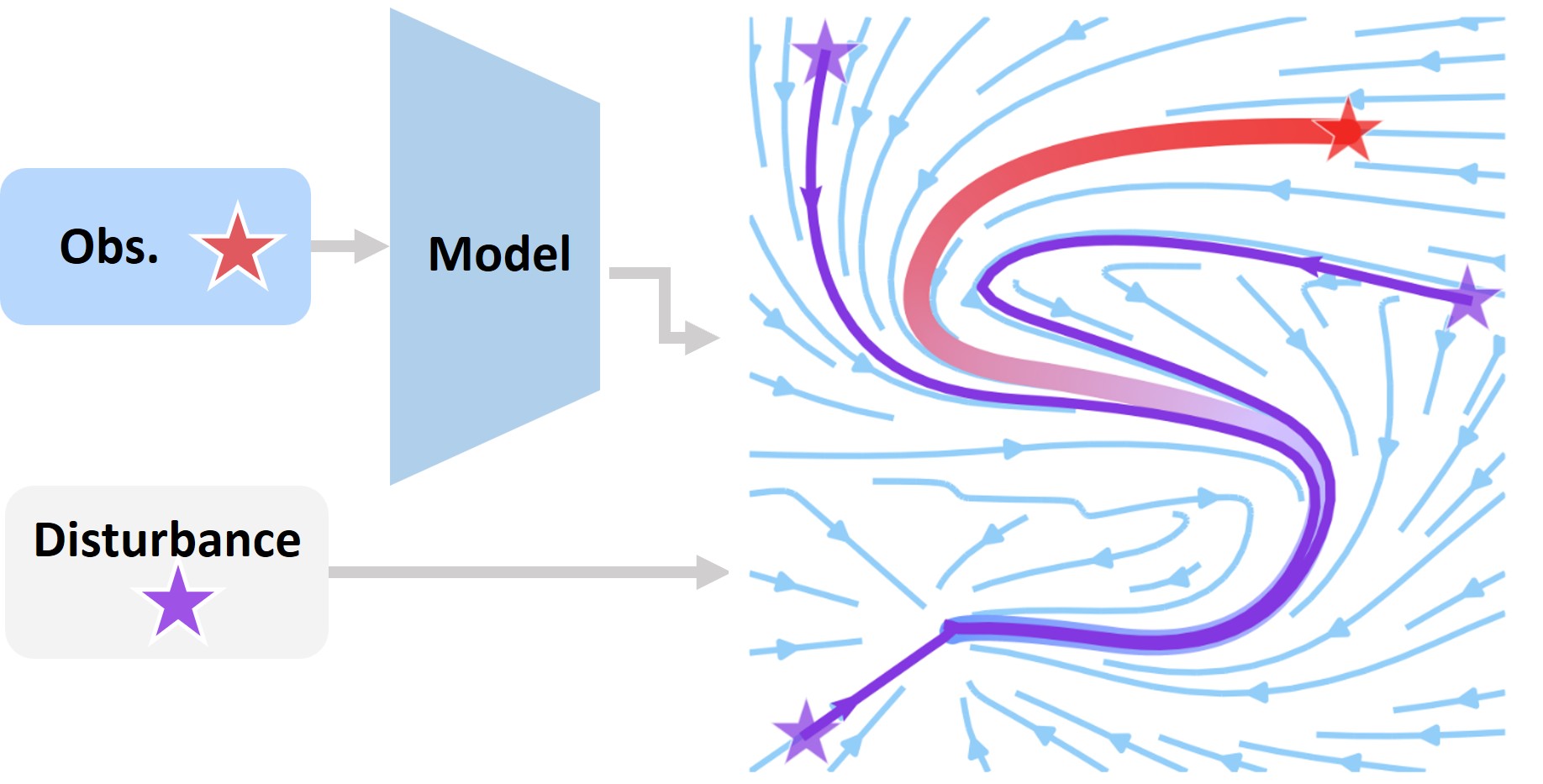}
    \caption{
    Perturbation recovery from the induced flow.
    The policy predicts a spline from the observation, and the spline-to-field transformation constructs the corresponding flow field.
    When disturbances move the system away from the nominal motion, the perturbed states enter the flow field and are guided back toward the spline geometry.
    }
    \label{fig:robust_generalization}
    \vspace{-10pt}
\end{figure}

A key motivation for the flow field realization is to provide state-dependent correction around the predicted motion. 
When execution is perturbed away from a nominal time-indexed trajectory, directly replaying the trajectory does not define how the robot should move from the perturbed state. 
In contrast, the induced flow field assigns a velocity to nearby states and can guide the system back toward the generated curve geometry, as illustrated in Fig.~\ref{fig:robust_generalization}.

This corrective behavior arises from the attraction--progression decomposition in \eqref{eq:attraction_progression}. 
At regular points, the attraction term decreases the distance to the spline. 
For an interior closest-point projection, the normal direction is orthogonal to the spline tangent, so the progression term is locally tangent and does not increase the distance to the spline. 
Therefore, the induced dynamics do not increase the distance to the generated spline~\cite{li2025movement}.

% The terminal point of the generated spline,
% \begin{equation}
%     \mathbf{x}^{\star}
%     =
%     \mathbf{f}_\theta(T),
% \end{equation}
% is made an equilibrium by imposing a zero terminal tangent:
% \begin{equation}
%     \dot{\mathbf{f}}_\theta(T)=\mathbf{0},
%     \qquad
%     \mathbf{F}_\theta(\mathbf{x}^{\star})=\mathbf{0}.
% \end{equation}

A derivation based on the distance-field Lyapunov function is provided in App.~\ref{app:flow_field_stability}. The policy determines which spline is generated from the observation. The flow-field realization can improve recovery from state perturbations around that predicted motion, but it does not by itself guarantee task success when the predicted spline is inappropriate for the task.

% Thus, the same predicted spline supports both time-indexed decoding and state-dependent flow-field execution.
% If the policy represents multiple distinct spline modes, applying $\mathcal{T}$ to each mode yields distinct flow fields.
% In practice, such multimodality can be handled by sampling or mode selection before the spline-to-field transformation.
% Averaging incompatible modes is not required by the formulation and may be undesirable for strongly multimodal behaviors.
% \vspace{-10pt}
\subsection{Uncertainty Propagation in Flow Fields}

The flow-field realization also supports uncertainty propagation, extending the trajectory-level formulation in Sec.~\ref{sec:uncertainty_spline}. 
Observation uncertainty first affects the predicted spline parameters $\mathbf{w}_\theta(\mathbf{o})$ and the decoded spline $\mathbf{f}_\theta$. 
The induced flow field then varies accordingly because the projection, distance, normal direction, and tangential progression are all computed from the predicted spline. 
Since this mapping from observations to flow vectors is generally nonlinear, we estimate the resulting flow-field distribution using Monte Carlo sampling.

For any state $\mathbf{x}$, the sampled flow vectors follow a Gaussian distribution
\begin{equation}
    \mathbf{F}_\theta(\mathbf{x}) \sim \mathcal{N}\bigl(
    \boldsymbol{\mu}_{\mathbf{F}}(\mathbf{x}),\;
    \boldsymbol{\Sigma}_{\mathbf{F}}(\mathbf{x})
    \bigr),
    \label{eq:flow_distribution}
\end{equation}
with $\boldsymbol{\mu}_{\mathbf{F}}(\mathbf{x})$ and $\boldsymbol{\Sigma}_{\mathbf{F}}(\mathbf{x})$ the mean and covariance of the flow vectors at state $\mathbf{x}$.

\begin{figure}[htbp]
    \centering
    \includegraphics[width=\linewidth]{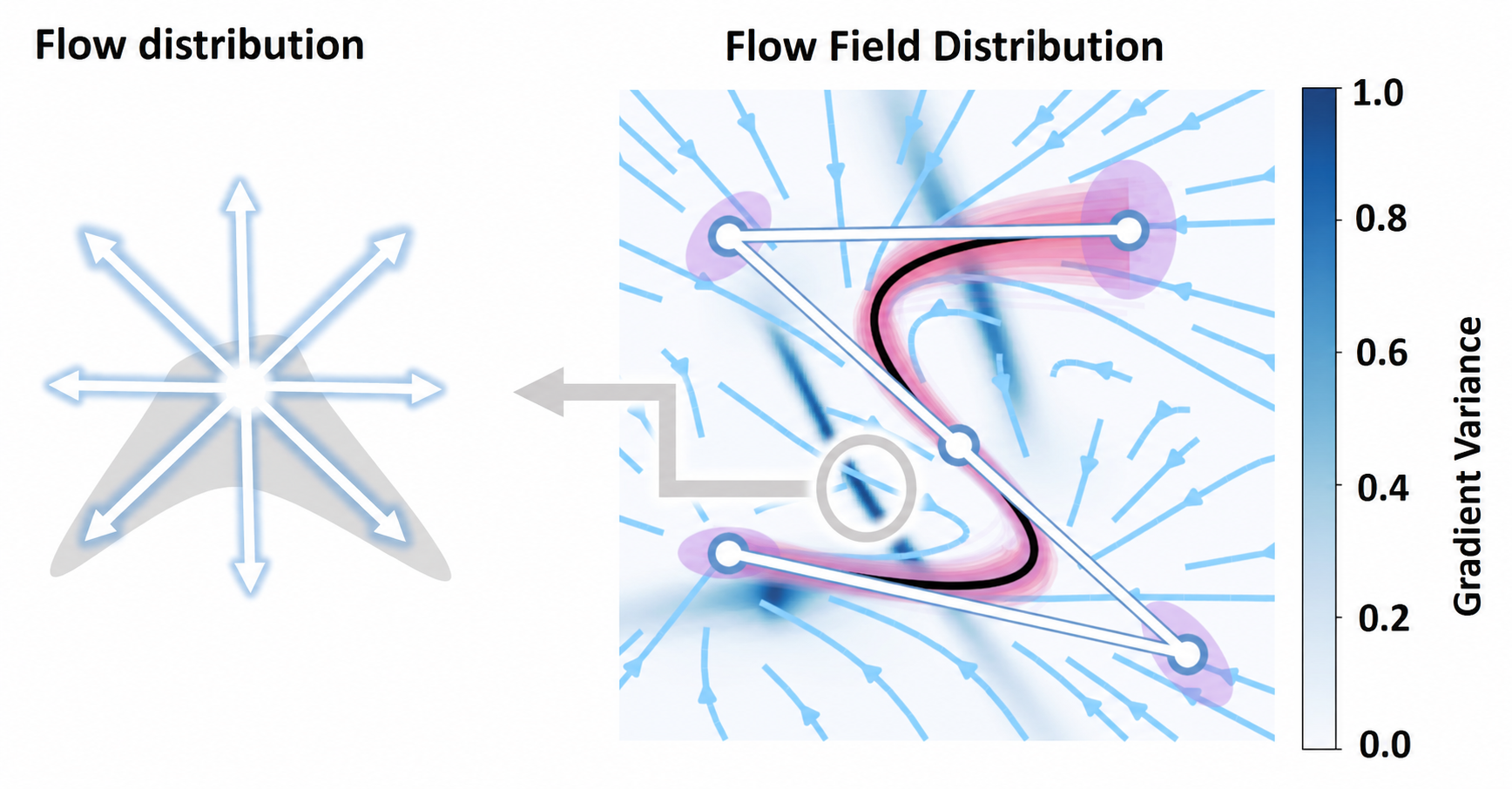}
    \vspace{-20pt}
    \caption{
    Flow-field uncertainty propagation.
    \textbf{Left:} Illustration of the flow-vector distribution.
    \textbf{Right:} Flow field induced by the trajectory distribution.
    The blue streamlines denote the mean flow field, while the color map shows the normalized flow-vector variance at each location.
    Darker regions indicate higher variance.
    }
    \label{fig:flow_uncertainty}
    \vspace{-10pt}
\end{figure}

This probabilistic formulation provides a spatial view of how observation uncertainty affects the induced dynamics. 
Under noisy observations, different samples may generate different splines and, therefore, different flow fields. 
The mean flow summarizes the average corrective direction, while the covariance reflects how consistently this direction is induced across samples. 
As visualized in Fig.~\ref{fig:flow_uncertainty}, regions with higher variation indicate that the local flow direction is more sensitive to observation uncertainty, whereas regions with lower variation correspond to more consistent induced dynamics. 
This provides an interpretable state-space uncertainty measure that complements the trajectory-level uncertainty described earlier.

In practice, we can estimate the flow-field distribution via Monte Carlo sampling. 
We draw samples $\{\mathbf{o}^{(m)}\}_{m=1}^{M}$ from the observation distribution and compute the corresponding spline parameters $\mathbf{w}_\theta(\mathbf{o}^{(m)})$ and decoded splines $\mathbf{f}^{(m)}_\theta$. 
Each sampled spline induces a flow field using the attraction--progression construction in \eqref{eq:attraction_progression} for the sampled spline $\mathbf{f}^{(m)}_\theta$.

%namely
% \begin{equation}
% \begin{aligned}
%     \mathbf{F}_\theta^{(m)}(\mathbf{x})
%     &=
%     -
%     \alpha(d_\theta^{(m)}(\mathbf{x}))\,
%     \mathbf{n}_\theta^{(m)}(\mathbf{x})
%     \\
%     &\quad+
%     \beta(d_\theta^{(m)}(\mathbf{x}))\,
%     \dot{\mathbf{f}}_\theta^{(m)}
%     \bigl(t_\theta^{(m)}(\mathbf{x})\bigr),
% \end{aligned}
%     \label{eq:flow_sample}
% \end{equation}
% where $d_\theta^{(m)}$, $\mathbf{n}_\theta^{(m)}$, and $t_\theta^{(m)}$ are computed from the sampled spline $\mathbf{f}^{(m)}_\theta$.

The mean and covariance of the flow field are then approximated as
\begin{equation}
    \boldsymbol{\mu}_{\mathbf{F}}(\mathbf{x})
    =
    \frac{1}{M}
    \sum_{m=1}^{M}
    \mathbf{F}_\theta^{(m)}(\mathbf{x}),
    \label{eq:flow_mean}
\end{equation}
\begin{equation}
    \boldsymbol{\Sigma}_{\mathbf{F}}(\mathbf{x})
    =
    \frac{1}{M}
    \sum_{m=1}^{M}
    \left(
    \mathbf{F}_\theta^{(m)}(\mathbf{x})
    -
    \boldsymbol{\mu}_{\mathbf{F}}(\mathbf{x})
    \right)
    \left(
    \mathbf{F}_\theta^{(m)}(\mathbf{x})
    -
    \boldsymbol{\mu}_{\mathbf{F}}(\mathbf{x})
    \right)^\top .
    \label{eq:flow_covariance}
\end{equation}
This procedure can be applied to a batch of query states to estimate flow-field uncertainty across the state space and can be efficiently parallelized.
\subsection{Integration with Robot Control}
\label{sec:flow_control}

The flow-field realization converts the spline output from a time-indexed trajectory into a state-dependent vector field around the generated motion. 
This makes the learned motion more suitable for closed-loop composition with execution-time objectives, such as obstacle avoidance or task constraints. 
We illustrate this interface with a specific task-priority formulation, where the SP flow provides the nominal manipulation behavior and collision avoidance is imposed as a higher-priority objective.

We consider an operational-space SP flow and map it to configuration space. 
Let $\mathbf{x}=\boldsymbol{\psi}(\mathbf{q})$ denote the operational-space task variables, with Jacobian $\mathbf{J}_\psi(\mathbf{q})$. 
The corresponding configuration-space velocity is
\begin{equation}
    \dot{\mathbf{q}}_{\theta}
    =
    \mathbf{J}_\psi^\dagger(\mathbf{q})\,
    \mathbf{F}_\theta(\boldsymbol{\psi}(\mathbf{q})),
    \label{eq:flow_to_configuration}
\end{equation}
where $\mathbf{J}_\psi^\dagger(\mathbf{q})$ is the Moore--Penrose pseudo-inverse. 

For collision avoidance, we use the robot's signed distance field $\Gamma_{\mathrm{SDF}}(\mathbf{x},\mathbf{q})$~\cite{li2024representing}. 
The high-priority avoidance velocity is defined as
\begin{equation}
    \dot{\mathbf{q}}_{\mathrm{col}}
    =
    \rho(\mathbf{x},\mathbf{q}) \;
    \nabla_{\mathbf{q}}
    \Gamma_{\mathrm{SDF}}(\mathbf{x},\mathbf{q})^\top,
    \label{eq:collision_velocity}
\end{equation}
where $\rho(\mathbf{x},\mathbf{q})$ activates and scales the repulsive motion according to the obstacle distance~\cite{simmoteit2025diffeomorphic}. 
When avoidance is active, the SP velocity is then projected into the null space of this avoidance direction with
\begin{equation}
    \dot{\mathbf{q}}_{\theta,\mathrm{proj}}
    =
    \left(
    \mathbf{I}
    -
    {\nabla_{\mathbf{q}}
    \Gamma_{\mathrm{SDF}}}^\dagger \; \nabla_{\mathbf{q}}
    \Gamma_{\mathrm{SDF}}
    \right)
    \dot{\mathbf{q}}_{\theta}.
    \label{eq:null_space_projection}
\end{equation}
The final command is
\begin{equation}
    \dot{\mathbf{q}}_{\mathrm{action}}
    =
    \dot{\mathbf{q}}_{\mathrm{col}}
    +
    \dot{\mathbf{q}}_{\theta,\mathrm{proj}}.
    \label{eq:final_control_policy}
\end{equation}

When no obstacle is present, $\rho$ is inactive, and the robot follows the SP flow. 
When the robot approaches an obstacle, the SDF term takes priority, while the learned SP motion is retained only in directions that do not interfere with avoidance to first order. 
This example shows how SP can provide a control-compatible nominal motion representation without retraining the policy backbone.

\begin{figure}[htbp]
    \centering
    \includegraphics[width=\linewidth]{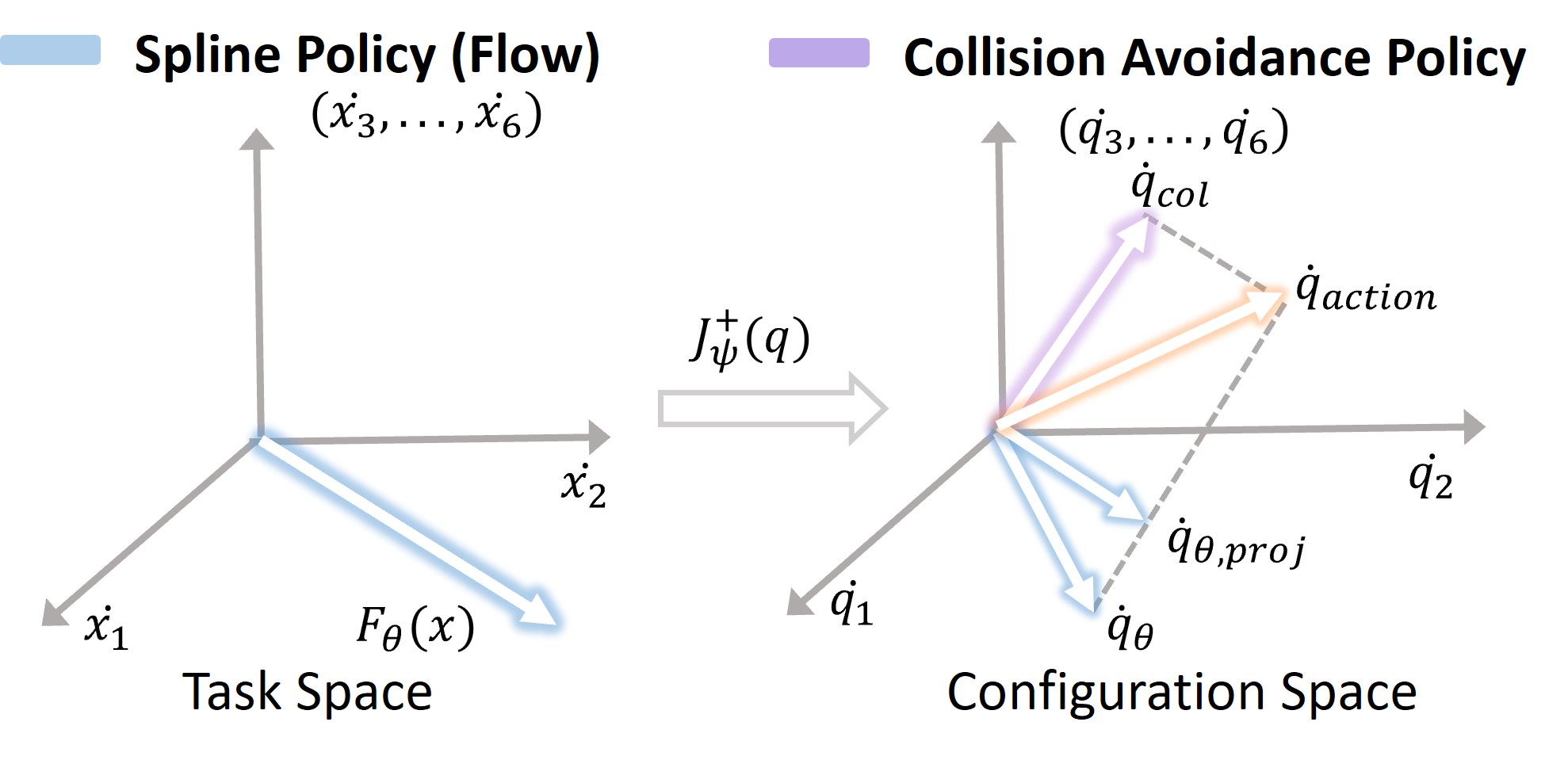}
    \vspace{-20pt}
    \caption{
    Null-space projection.
    The final command $\dot{\mathbf{q}}_{\mathrm{action}}$ combines a high-priority collision-avoidance velocity $\dot{\mathbf{q}}_{\mathrm{col}}$ with the SP velocity projected into the null space of the avoidance direction.
    }
    \label{fig:null_space_mechanism}
\end{figure}

\vspace{-15pt}
\section{Results}

We organize the experiments according to the role of SP as a structured representation, moving from low-dimensional studies to simulated manipulation benchmarks to real-world studies.
In low-dimensional settings, we first study the flow-field realization, where perturbation recovery and uncertainty effects can be directly visualized and quantified.
We then evaluate the spline trajectory representation on simulated manipulation benchmarks under matched policy backbones.
Finally, we present real-robot case studies illustrating how the same spline output can be connected to visual replanning, disturbance recovery, null-space collision avoidance, externally specified motions, and different policy backbones.

\subsection{Flow Field of Spline Policy}

\textbf{Experimental setup:}
We begin with the LASA dataset~\cite{khansari2011learning}, a standard low-dimensional handwriting-motion benchmark widely used in learning-from-demonstration studies.
% Since the original LASA dataset provides demonstrations without observations, we construct a simplified observation-conditioned setting by using the initial position of each demonstration as the observation.
% This setting is used as a controlled mechanism study rather than as a full perception-conditioned manipulation benchmark.
We evaluate two aspects of the flow-field realization.
First, we introduce perturbations to the state to evaluate local correction and endpoint convergence.
Second, we inject Gaussian noise into the observations to evaluate consistency under observation uncertainty.

\begin{figure}[htbp]
    \centering
    \includegraphics[width=0.9\linewidth]{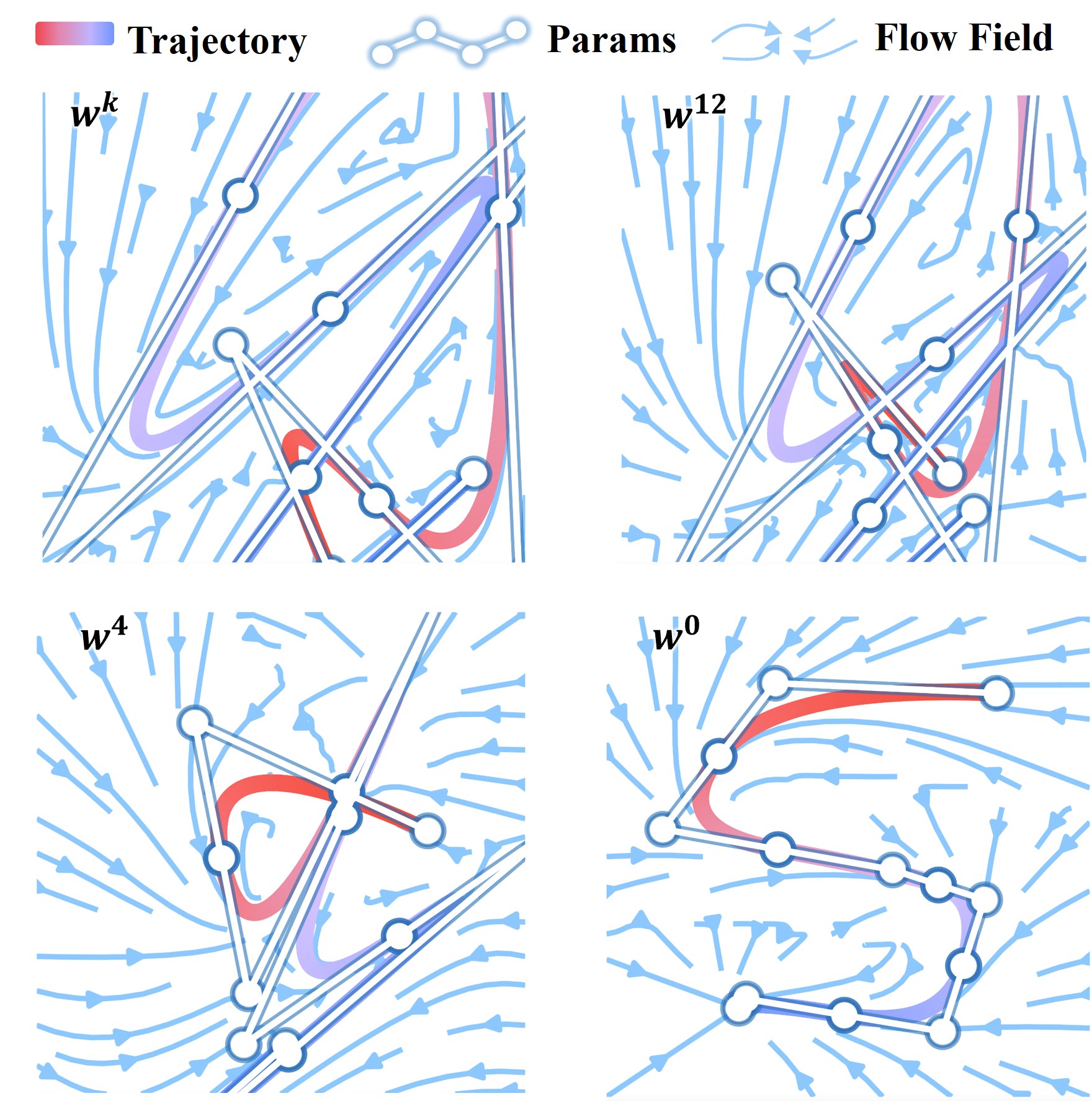}
    \caption{
    Generation of SP.
    For visualization, we use diffusion as the base model, whose iterative denoising process illustrates the synchronous evolution of spline parameters, trajectories, and the flow field from $\mathbf{w}^{k}$ to $\mathbf{w}^{0}$.
    }
    \label{fig:denoising_dds}
    \vspace{-10pt}
\end{figure}

\textbf{Methods:}
We use Diffusion Policy~\cite{chi2024diffusionijrr} as the baseline in this study.
The \textit{Baseline model} follows the standard DP formulation and directly maps observations to actions.
\textit{Spline Policy (Traj.) (ours)} predicts spline parameters and executes the decoded trajectory.
\textit{Spline Policy (Flow) (ours)} converts the predicted spline into the corresponding flow field.
To study uncertainty propagation, \textit{Spline Policy (Prob. Traj.) (ours)} estimates a distribution over spline parameters, while \textit{Spline Policy (Prob. Flow) (ours)} estimates the induced distribution over flow fields.
We first visualize the learned SP in Fig.~\ref{fig:denoising_dds}, followed by quantitative evaluations.

\textbf{Evaluation metrics:}
We report three metrics.
One-way Chamfer Distance (CD) from the generated trajectory to the target trajectory measures how closely the generated trajectory matches the target.
Convergence Error (CE) measures the deviation of the final state from the target endpoint.
Maximum Speed (MS) reports the maximum speed along the generated motion and is used as a simple proxy for abrupt motions.

\begin{figure}[!h]
    \centering
    \includegraphics[width=\linewidth]{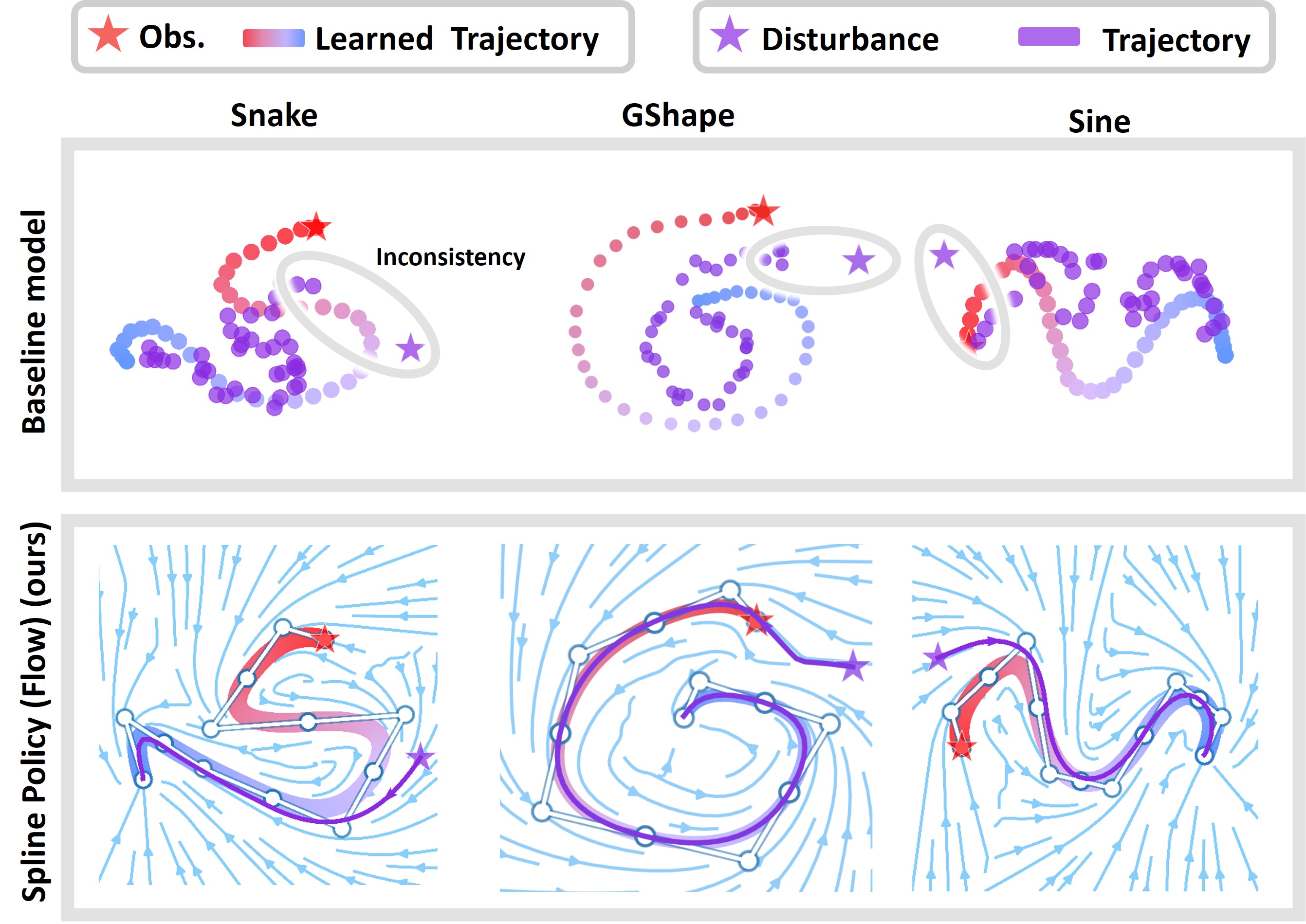}
    \vspace{-15pt}
    \caption{
    Comparison between \textbf{Baseline model} and \textbf{Spline Policy (Flow) (ours)} across three kinds of demonstrations: Snake, G-shape, and Sine.
    The purple observations indicate perturbed states.
    The gray ellipses highlight that the baseline model breaks spatial consistency under perturbations.
    }
    \label{fig:mechanism_bench}
    \vspace{-5pt}
\end{figure}

\textbf{Perturbation recovery:}
As shown in Fig.~\ref{fig:mechanism_bench}, Spline Policy (Flow) (ours) guides perturbed states back toward the demonstrated motion through the induced state-dependent vector field.
Under this execution protocol, the trajectory-output variants do not define a corrective direction away from the predicted curve unless an additional feedback controller or replanning mechanism is introduced.
Replanning from perturbed states would also require additional policy queries and depend on whether such states are covered by the training distribution.
We repeat the experiment with 25 perturbations for each of the three demonstration types, Snake, G-Shape, and Sine, and summarize the average results in Table \ref{tab:lasa_dataset_exp}.
Spline Policy (Flow) (ours) achieves the lowest values across the reported metrics in this setting, indicating lower trajectory error, lower endpoint error, and lower maximum speed under the tested perturbations.

\begin{table}[t]
\caption{Quantitative evaluation under disturbances.}
\label{tab:lasa_dataset_exp}
\centering
\begin{tabular}{lcccc}
\toprule
Method 
& \makecell{CD (mean) \\ {[mm]}}
& \makecell{CD (min) \\ {[mm]}}
& \makecell{CE \\ {[mm]}}
& \makecell{MS \\ {[m/s]}} \\
\midrule
Baseline model
& 26.2
& 3.7
& 26.7
& 13.0 \\

Spline Policy (Traj.) (ours)
& 21.0
& 2.2
& 28.8 
& 14.3 \\

Spline Policy (Flow) (ours)
& \textbf{12.8} 
& \textbf{0.8} 
& \textbf{1.1} 
& \textbf{0.25} \\
\bottomrule
\end{tabular}
\end{table}

\textbf{Observation uncertainty:}
We next inject Gaussian noise into the observations.
The Baseline, Spline Policy (Traj.), and Spline Policy (Flow) directly take noisy observations as input.
Spline Policy (Prob. Traj.) and Spline Policy (Prob. Flow) instead estimate distributions over spline parameters and flow fields, respectively, and use the distribution mean to generate the final output.
The results are reported in Table \ref{tab:obs_noise_uncertainty}.
As the noise level increases, the non-probabilistic variants become more sensitive to observation noise.
In this injected-noise protocol, the probabilistic variants achieve lower Chamfer distances across all tested noise levels, suggesting that Monte Carlo propagation through the spline representation provides a useful averaging mechanism.

\begin{table}[t]
\caption{Chamfer distance (CD) under observation uncertainty.}
\label{tab:obs_noise_uncertainty}
\centering
\begin{tabular}{lcccc}
\toprule
Method
& \makecell{$\sigma$ = 10 \\ {[mm]}}
& \makecell{$\sigma$ = 20 \\ {[mm]}}
& \makecell{$\sigma$ = 30 \\ {[mm]}}
& \makecell{$\sigma$ = 40 \\ {[mm]}} \\
\midrule
Baseline model
& 8.6
& 14.3
& 18.0
& 20.4 \\

\makecell[l]{Spline Policy (Traj.)(ours)}
& 4.5
& 8.7
& 12.4
& 15.2 \\

\makecell[l]{Spline Policy (Flow)(ours)}
& 2.7
& 6.5
& 14.6
& 26.6 \\

\makecell[l]{Spline Policy (Prob. Traj.)(ours)}
& \textbf{\textcolor{bestblue}{0.7}}
& \textbf{\textcolor{bestblue}{2.1}}
& \textbf{\textcolor{bestblue}{4.0}}
& \textbf{\textcolor{bestblue}{6.1}} \\

\makecell[l]{Spline Policy (Prob. Flow)(ours)}
& \textbf{\textcolor{secondblue}{1.3}}
& \textbf{\textcolor{secondblue}{2.7}}
& \textbf{\textcolor{secondblue}{4.7}}
& \textbf{\textcolor{secondblue}{7.2}} \\
\bottomrule
\end{tabular}

\vspace{2pt}
{\footnotesize
Best results are highlighted in bold blue, and second-best results are shown in light blue.
$\sigma$: standard deviation of Gaussian noise applied to observations.
}
\vspace{-5pt}
\end{table}

\begin{figure}[!h]
    \centering
    \includegraphics[width=0.9\linewidth]{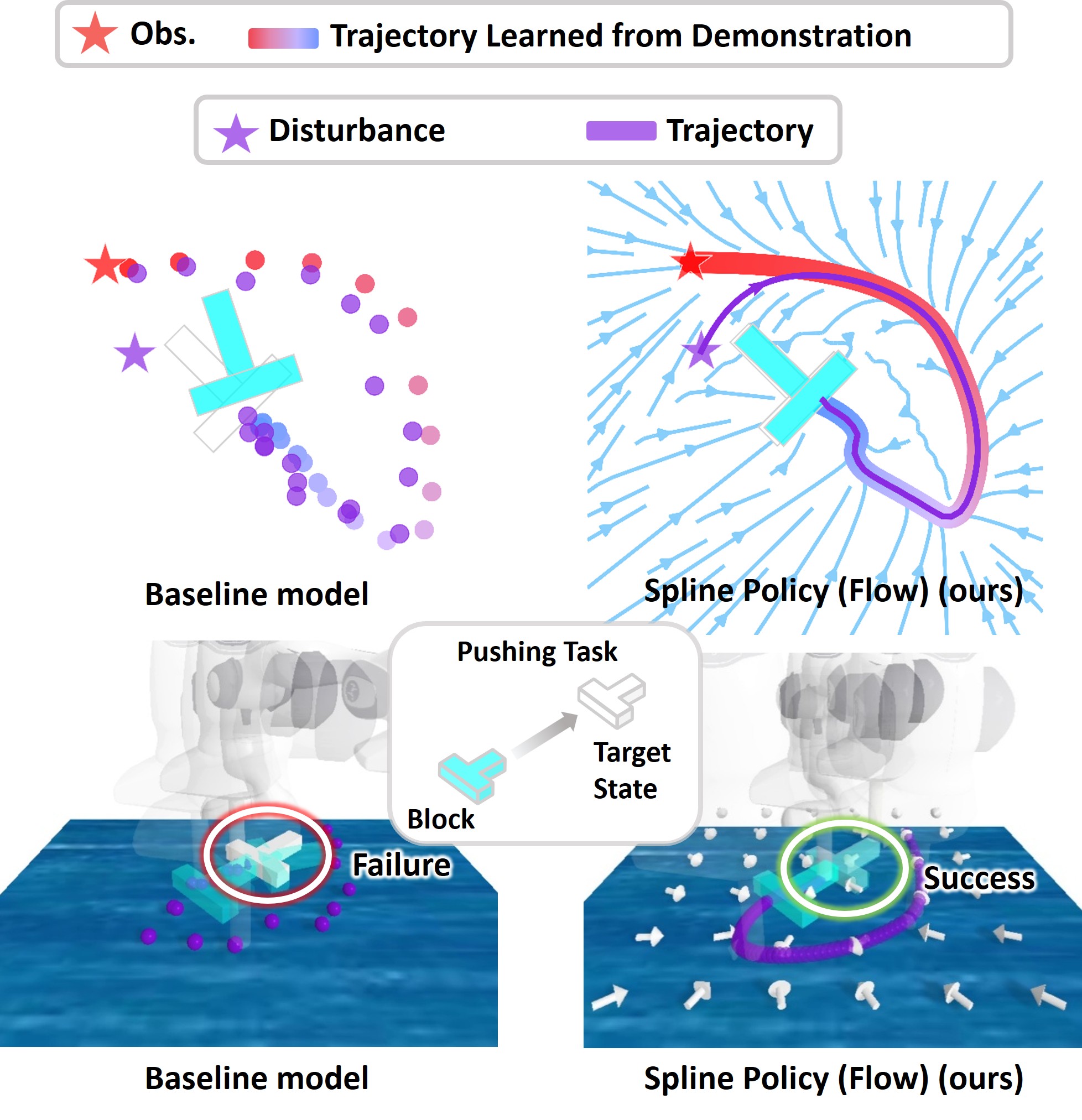}
    \vspace{-10pt}
    \caption{
    Qualitative pushing illustration of \textbf{Baseline model} and \textbf{Spline Policy (Flow) (ours)} under perturbed initial states.
    }
    \label{fig:generalization_bench}
    \vspace{-10pt}
\end{figure}

\textbf{Qualitative contact-rich illustration:}
We further illustrate the flow-field realization in a contact-rich pushing task, where the robot pushes a block toward a target position using a limited number of demonstrations~\cite{florence2022implicit}.
At each trial, the initial position of each demonstration is used as the input state.
As shown in Fig.~\ref{fig:generalization_bench}, this example illustrates how the induced flow can guide recovery from initial-state variations in the tested setup.

\subsection{Representation of Spline Policy}

We evaluate SP as a trajectory-output representation.
The goal of this benchmark is to test whether replacing a fixed-resolution action chunk with a compact spline output keeps task performance in a comparable range under matched policy backbones, while reducing policy-output dimensionality and network-level forward FLOPs.
We evaluate this representation across state, vision, and point-cloud inputs, including high-dimensional and dexterous manipulation tasks.

\textbf{Tasks:}
We evaluate six robotic manipulation tasks covering three input modalities.
The state-based transport (PH) task requires precise multistage motion.
The vision tasks Push-T~\cite{florence2022implicit} and Can (PH)~\cite{mandlekar2021matters} evaluate contact-rich manipulation and goal-directed placement from RGB images.
With 3D point-cloud input, Dexart Laptop~\cite{bao2023dexart} assesses articulated-object handling, whereas Adroit Door and Adroit Pen~\cite{rajeswaran2018learning} require dexterous manipulation skills.
These tasks are implemented in MuJoCo~\cite{todorov2012mujoco}, Sapien~\cite{xiang2020sapien}, and Pymunk~\cite{pymunk}, covering multiple simulator environments.

\textbf{Perception encoder:}
For state input, we directly feed the low-dimensional state vector into the network.
For vision input, we use ResNet-18~\cite{he2016deep} to extract image features from RGB observations.
For 3D point-cloud input, we use the lightweight MLP encoder from~\cite{ze2024dp3}.

\textbf{Methods:}
We use DP~\cite{chi2024diffusionijrr} and FMP~\cite{zhang2024affordance} as action-chunk baseline models, denoted as BL-Diff and BL-Flow.
Using the same backbones, we implement Spline Policy (Traj.) (ours) with diffusion and flow-matching objectives, denoted as SP-Diff (ours) and SP-Flow (ours).
The action-chunk representation $\mathbb{R}^{16 \times d_a}$ is replaced by a spline-parameter representation $\mathbb{R}^{8 \times d_a}$, where $d_a$ is the action dimension, $16$ is the action horizon, and $8$ is the number of spline parameters per action dimension used by the decoder.
For each matched-backbone comparison, the corresponding methods use the same perception encoder and training settings.

\textbf{Evaluation metrics:}
We evaluate the four methods on each task $T_i$ using the task score and relative Floating Point Operations (FLOPs).
For each method, the score is reported as the maximum and the average over the best five checkpoints.
The FLOP numbers measure the policy-network forward pass only; spline decoding and execution-mode-specific costs are not included.
To measure relative network-level cost, we normalize the FLOPs of method $m$ by the corresponding action-chunk baseline $\pi(m)$:
\begin{equation}
\pi(m) =
\begin{cases}
\text{BL-Diff}, & m \in \{\text{BL-Diff},\, \text{SP-Diff (ours)}\}, \\[4pt]
\text{BL-Flow}, & m \in \{\text{BL-Flow},\, \text{SP-Flow (ours)}\}.
\end{cases}
\end{equation}
The relative FLOPs are then
\begin{equation}
\rho_{m}(T_i) = \frac{\text{FLOPs}_{m}(T_i)}{\text{FLOPs}_{\pi(m)}(T_i)}.
\end{equation}

\begin{figure}[!h]
    \includegraphics[width=\linewidth]{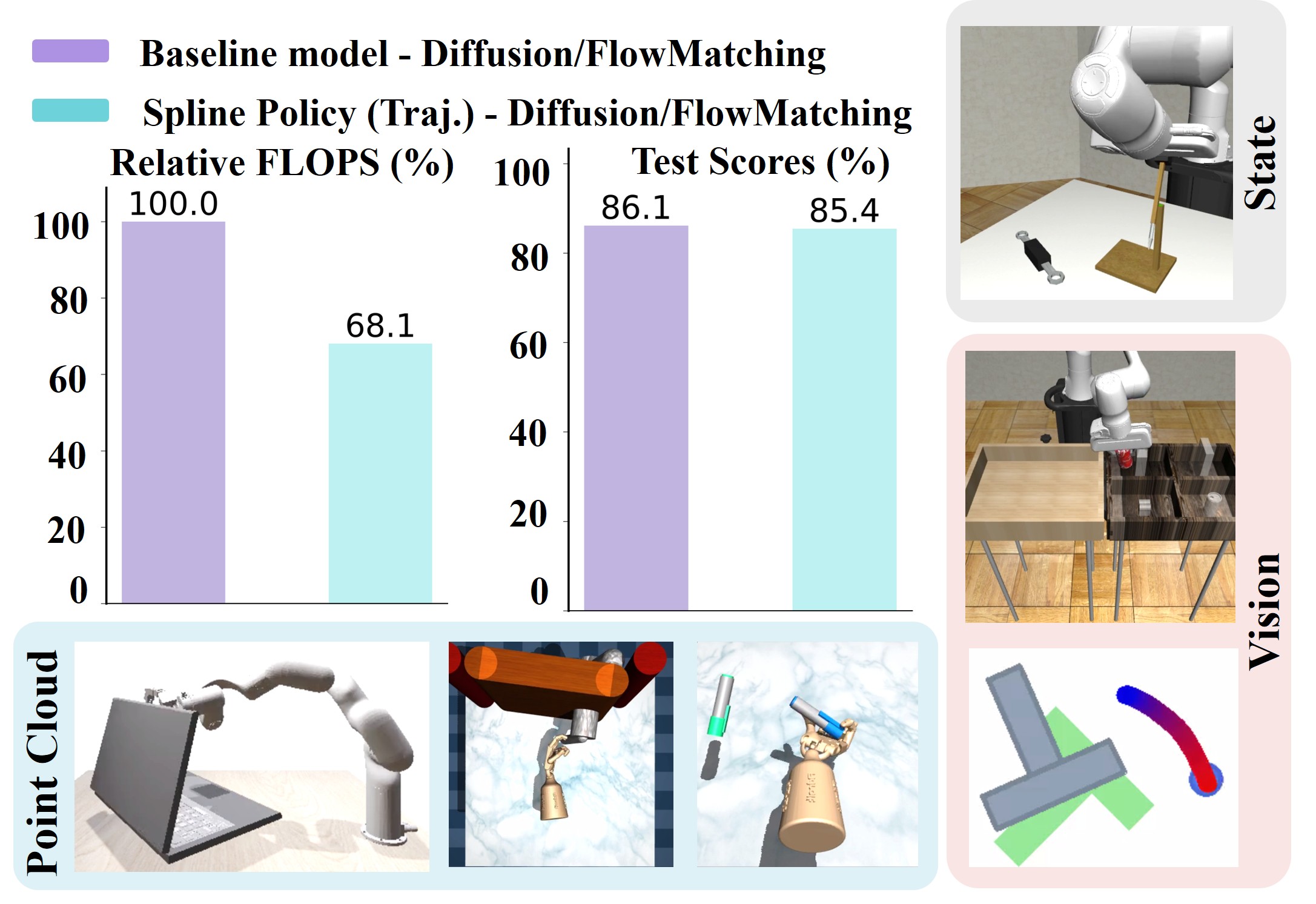}
    \vspace{-15pt}
    \caption{
    Benchmark between \textbf{Baseline model} and \textbf{Spline Policy (Traj.) (ours)} on tasks spanning state, vision, and point-cloud inputs.
    Test scores and network-level relative FLOPs are averaged over six tasks.
    % The spline output reduces policy-network forward FLOPs in this matched setup, while task scores remain close to the corresponding action-chunk baselines with task-dependent gains and losses.
    }
    \label{fig:benchmark}
    % \vspace{-10pt}
\end{figure}

\textbf{Evaluation results:}
The results in Table \ref{tab:bc_benchmark_singlecol} show task-dependent differences between action-chunk and spline-output policies.
SP matches or slightly improves some entries, while trailing the corresponding action-chunk baseline in others.
The consistent effect is the reduction in policy-output dimensionality and measured network-level forward FLOPs.
As summarized in Fig.~\ref{fig:benchmark}, the averaged scores remain close to the baselines, while the averaged relative FLOPs are reduced.
We therefore interpret this benchmark as evidence that the spline-output replacement remains in a comparable performance range under matched backbones, rather than as a claim of uniform performance improvement.

\begin{table}[!h]
\caption{Behavior cloning benchmark.}
\label{tab:bc_benchmark_singlecol}
\centering
\begin{tabular}{l|c|cc cc}
\toprule
\textbf{Task} & \textbf{Metric} & \makecell{\textbf{BL-Diff}} & \makecell{\textbf{BL-Flow}}
& \makecell{\textbf{SP-Diff}\\(ours)}
& \makecell{\textbf{SP-Flow}\\(ours)} \\
\midrule

\multicolumn{6}{c}{\textbf{State}} \\

\multirow{2}{*}{\makecell[l]{Transport}}
& Score 
& \textbf{0.88}/\textbf{0.80} 
& 0.84/\textbf{0.80} 
& 0.80/0.76 
& \textbf{0.88}/0.78 \\

& FLOPs (\%)
& 100.00 & 100.00 & \textbf{50.67} & \textbf{50.67} \\

\midrule
\multicolumn{6}{c}{\textbf{Vision}} \\

\multirow{2}{*}{Can}
& Score 
& \textbf{1.0}/\textbf{1.0} 
& \textbf{1.0}/\textbf{1.0} 
& \textbf{1.0}/\textbf{1.0} 
& \textbf{1.0}/\textbf{1.0} \\

& FLOPs (\%)
& 100.00 & 100.00 & \textbf{87.63} & \textbf{87.63} \\
\addlinespace[4pt]

\multirow{2}{*}{Push-T}
& Score 
& \textbf{0.98}/\textbf{0.97} 
& 0.97/\textbf{0.97} 
& 0.95/0.94 
& 0.96/0.95 \\

& FLOPs (\%)
& 100.00 & 100.00 & \textbf{87.63} & \textbf{87.63} \\

\midrule
\multicolumn{6}{c}{\textbf{Point Cloud}} \\

\multirow{2}{*}{\makecell[l]{Adroit \\ Door}}
& Score 
& \textbf{0.95}/0.81 
& 0.80/0.78 
& \textbf{0.95}/0.87 
& \textbf{0.95}/\textbf{0.88} \\

& FLOPs (\%)
& 100.00 & 100.00 & \textbf{61.03} & \textbf{61.03} \\
\addlinespace[4pt]

\multirow{2}{*}{\makecell[l]{Adroit \\ Pen}}
& Score 
& 0.80/0.74 
& \textbf{0.85}/\textbf{0.75} 
& \textbf{0.85}/0.68 
& 0.80/\textbf{0.75} \\

& FLOPs (\%)
& 100.00 & 100.00 & \textbf{61.03} & \textbf{61.03} \\
\addlinespace[4pt]

\multirow{2}{*}{\makecell[l]{Dexart \\ Laptop}}
& Score 
& \textbf{0.95}/\textbf{0.87} 
& \textbf{0.95}/0.85 
& 0.90/0.81 
& 0.90/0.84 \\

& FLOPs (\%)
& 100.00 & 100.00 & \textbf{61.03} & \textbf{61.03} \\

\bottomrule
\end{tabular}

\vspace{2pt}
{\footnotesize
BL = Baseline, SP = Spline Policy (Traj.), Diff = Diffusion, Flow = Flow Matching.
Scores are reported as \textbf{max/average} over the best five checkpoints.
% Relative FLOPs are normalized with respect to the corresponding action-chunk baseline and measure the policy-network forward pass.
Diffusion and Flow Matching share identical architectures, resulting in identical relative FLOPs.
}
\vspace{-10pt}
\end{table}

\begin{figure}[!h]
    \centering
    \includegraphics[width=1.0\linewidth]{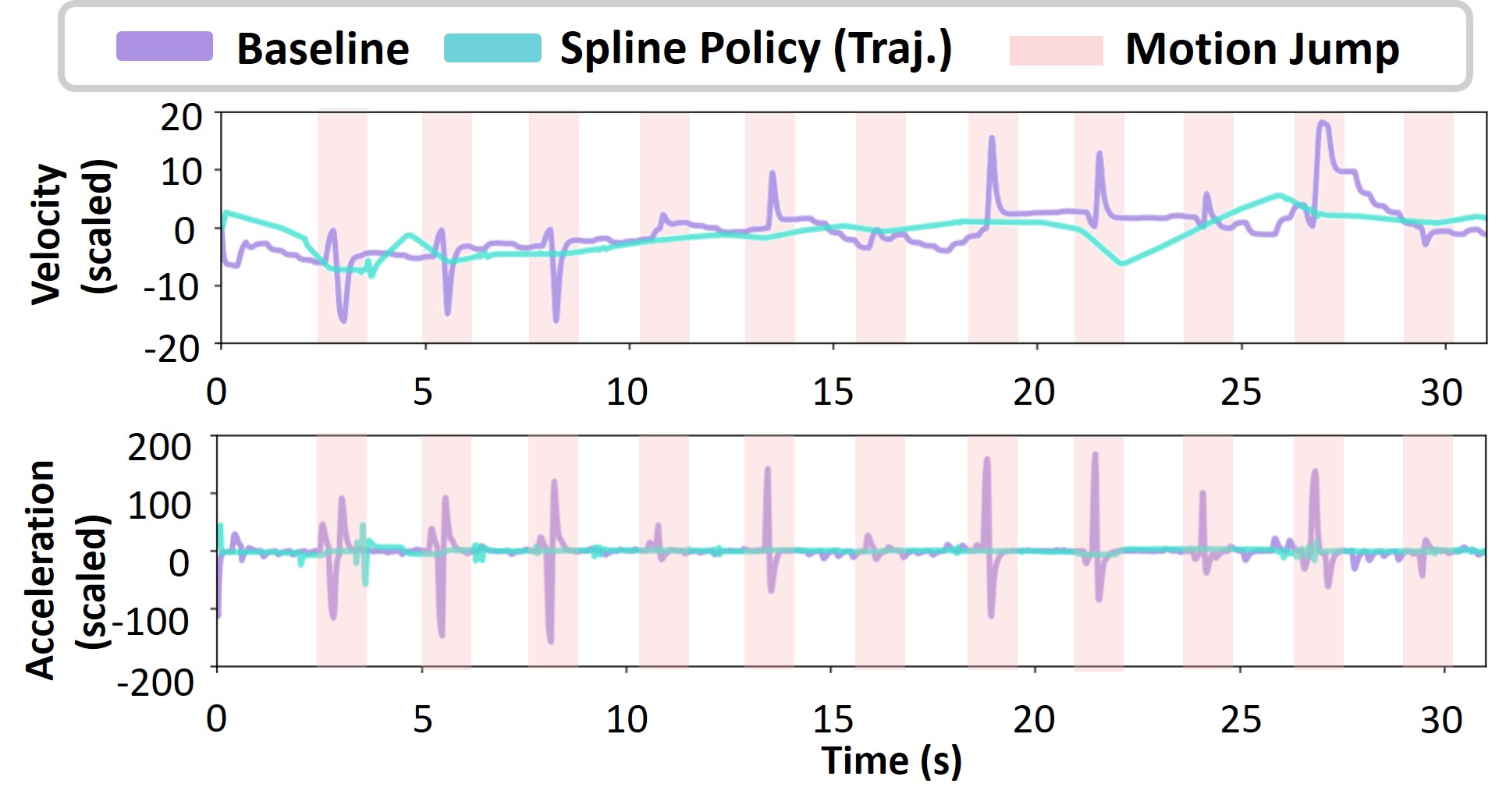}
    \vspace{-20pt}
    \caption{
    Trajectory continuity comparison between \textbf{Baseline model} and \textbf{Spline Policy (Traj.) (ours)} under replanning.
    }
    \label{fig:full_continuity}
    \vspace{-5pt}
\end{figure}

\textbf{Trajectory continuity under replanning:}
We further examine trajectory continuity on the Push-T~\cite{florence2022implicit} task.
As shown in Fig.~\ref{fig:full_continuity}, Spline Policy (Traj.) (ours) can impose $C^1$ continuity across replanned trajectories, resulting in smoother velocity profiles.
In contrast, the action-chunk baseline can exhibit velocity discontinuities at replanning boundaries.
Although acceleration continuity is not explicitly enforced in this experiment, the plotted acceleration profiles are less abrupt under the tested replanning setting.

\subsection{Real-World Compatibility Case Studies}

We present real-world case studies to illustrate how SP can be deployed with visual replanning, flow execution, null-space collision avoidance, and externally specified spline motions involving a liquid-filled object.
These studies are intended as system-level demonstrations of compatibility, rather than controlled comparisons against action-chunk policies.

\begin{figure*}[ht]
    \centering
    \includegraphics[width=1.0\linewidth]{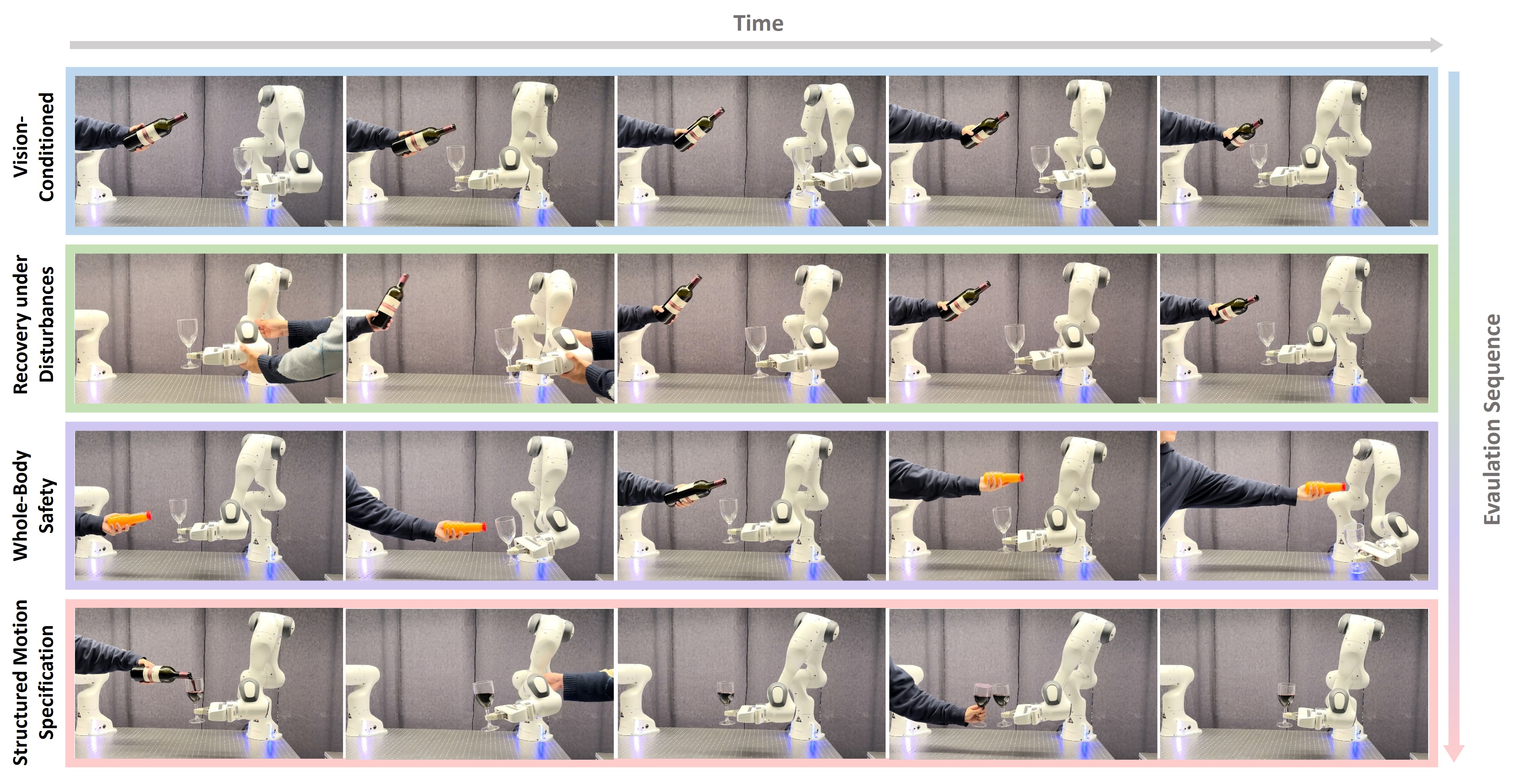}
    \vspace{-20pt}
    \caption{
    Real-world compatibility case studies.
    \textbf{Row 1:} Vision-conditioned planning under changing spatial configurations.
    \textbf{Row 2:} Recovery under repeated pushing and pulling disturbances.
    \textbf{Row 3:} Whole-body collision avoidance via null-space projection in the presence of an undesired object.
    \textbf{Row 4:} Structured motion specification and disturbance trials with a liquid-filled glass.
    }
    \label{fig:system_evaluation}
    \vspace{-20pt}
\end{figure*}

As illustrated in Fig.~\ref{fig:system_evaluation}, we organize the hardware demonstrations into four case studies:
\begin{enumerate}
    \item \textbf{Vision-conditioned planning}, which illustrates visual conditioning under changing spatial configurations.
    \item \textbf{Recovery under disturbances}, which illustrates qualitative recovery from pushing and pulling perturbations during execution.
    \item \textbf{Whole-body collision avoidance via null-space projection}, which illustrates how a collision-avoidance controller can be incorporated as a higher-priority behavior.
    \item \textbf{Structured motion specification}, which illustrates that the same spline representation can encode smooth task motions beyond learned policy outputs and remains compatible under disturbances with a liquid-filled glass.
\end{enumerate}

\begin{figure}[!h]
    \centering
    \includegraphics[width=1.0\linewidth]{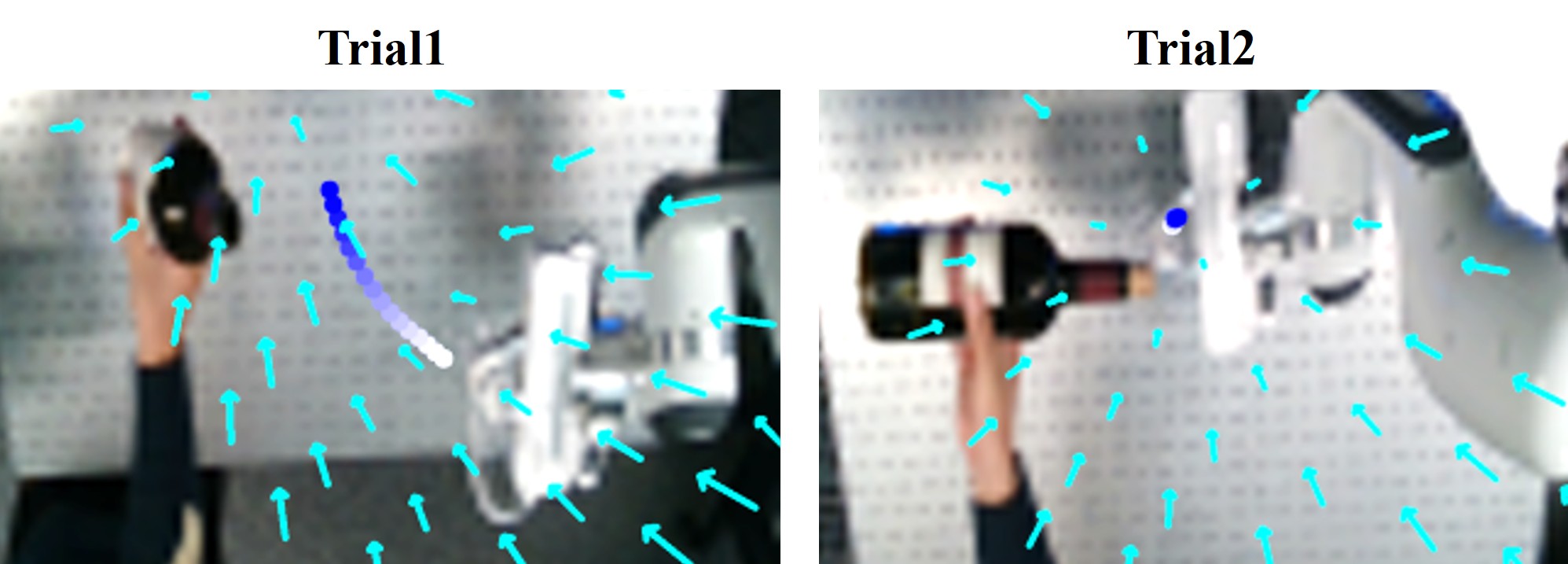}
    \caption{
    Vision-conditioned planning with Spline Policy (Flow).
    The cyan arrows indicate the induced flow field in the image, while the blue curve shows the Spline Policy (Traj.) representation over time, with the white segment highlighting the most recent motion.
    }
    \label{fig:real_flowfield}
    \vspace{-10pt}
\end{figure}

\textbf{Vision-conditioned planning:}
We first consider a vision-conditioned task in which a human holds a wine bottle at varying positions while the robot guides a wine glass toward it using visual observations (Fig.~\ref{fig:system_evaluation}, Row 1).
The policy is trained on 120 human demonstrations covering diverse relative configurations among the glass, bottle, and human hand.
The behavior is observed across different spatial arrangements, which is consistent with planning conditioned on visual observations rather than replaying a single fixed demonstration, as visualized in Fig.~\ref{fig:real_flowfield}.

\textbf{Recovery under disturbances:}
We next consider the system under external physical disturbances (Fig.~\ref{fig:system_evaluation}, Row 2).
While the robot is delivering the wine glass toward the bottle, a human repeatedly pushes or pulls the end-effector.
In these controlled trials, Spline Policy (Flow) generates motions that steer the end-effector back toward the generated motion.
This behavior is consistent with the combined effect of the state-dependent flow field and replanning from updated visual observations, as illustrated in Fig.~\ref{fig:real_flowfield}.

\textbf{Whole-body collision avoidance via null-space projection:}
We consider the system when an undesired object is introduced (Fig.~\ref{fig:system_evaluation}, Row 3).
With the SDF-derived collision-avoidance controller, the robot maintains clearance from the undesired object in the demonstrated scenario, including interactions involving the arm body.
Using null-space projection, the collision-avoidance command is assigned higher priority, while the SP flow is retained only in the remaining degrees of freedom.
This illustrates a training-free integration mechanism for collision avoidance, rather than a formal safety guarantee under arbitrary perception, modeling, latency, or actuation errors.

\textbf{Structured motion specification with disturbance trials:}
We also test whether the same spline representation can be used to specify smooth task motions outside the learned policy output (Fig.~\ref{fig:system_evaluation}, Row 4).
As an example, we encode a gentle shaking motion and execute it through the same downstream control pipeline, illustrating that the spline output object used by SP remains compatible with externally specified or edited motions when such a structure is useful for deployment. We additionally run controlled disturbance trials while the robot holds a liquid-filled glass, including sudden pulls and gentle contact interactions, as a stress test of real-world compatibility under controlled conditions; for safety, perturbation magnitudes were kept within limits.

\begin{figure*}[t]
    \centering
    \includegraphics[width=1.0\linewidth]{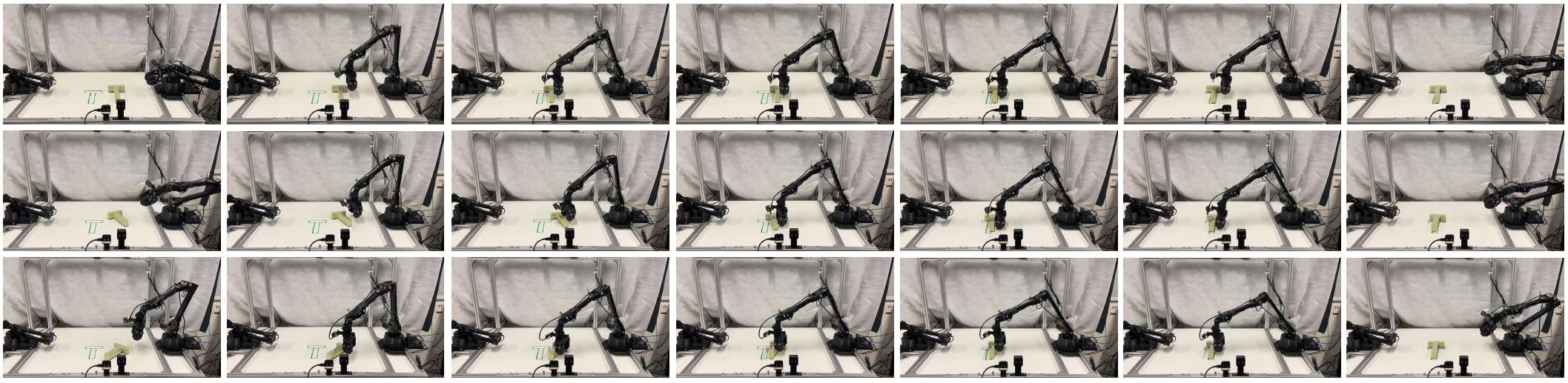}
    \vspace{-20pt}
    \caption{
    Spline Policy (Traj.) - PI05 on the PushT task under randomized initial positions.
    The language instruction is ``Push the T-shape block into a green T marker''.
    Left to right: snapshots of the ALOHA robot executing the task with different initial configurations of the T-block.
    }
    \label{fig:result_tblock}
    \vspace{-10pt}
\end{figure*}

\begin{figure*}[t]
    \centering
    \includegraphics[width=1.0\linewidth]{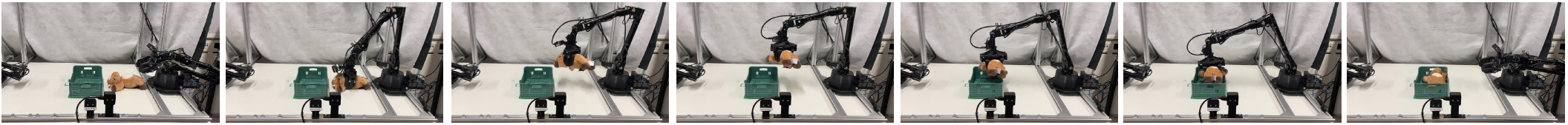}
    \vspace{-20pt}
    \caption{
    Spline Policy (Traj.) - PI05 on the toy packing task.
    The language instruction is ``pick and place a toy sheep into a green basket''.
    Left to right: snapshots of the ALOHA robot executing the task.
    }
    \label{fig:result_toy}
    \vspace{-10pt}
\end{figure*}

\subsection{Real-World Deployment with Different Backbones}

To further illustrate the model-agnostic nature of SP, we instantiate Spline Policy (Traj.) with different real-world policy backbones, including ACT, DP, and the VLA model PI05. 
In each case, the original policy backbone is preserved, while the trajectory-output representation is changed to spline parameters. 
The policies are evaluated on two ALOHA manipulation tasks, PushT and toy packing, under randomized object poses. 
The PI05-based spline policy rollouts are shown in Fig.~\ref{fig:result_tblock} and Fig.~\ref{fig:result_toy}. 
Detailed training settings, dataset statistics, and success rates are provided in App.~\ref{sec:realworld_backbone_details}. 
These experiments are intended to illustrate deployment with different backbones, rather than to provide a controlled comparison between backbone architectures, since model size, pretraining, optimizer, and training schedules differ across methods.

\section{Conclusion}

This paper studied whether the fixed-resolution action chunks commonly used in modern imitation-learning policies can be replaced by spline-parameter outputs while keeping the policy backbone unchanged.
The resulting Spline Policy (SP) formulation makes the predicted action object a compact continuous trajectory rather than a discrete sequence of points.
This output representation exposes temporal resampling, derivative access, boundary and continuity structure, and parameter-space operations before execution, while remaining compatible with diffusion, flow-matching, transformer, and VLA-style policy backbones.

For the quadratic spline construction considered in this work, the same predicted spline can also be converted into a state-dependent flow field through an analytic distance field construction.
This flow-field realization provides local correction around the generated motion: under the stated regularity and projection assumptions, the induced dynamics do not increase the distance to the predicted spline, and the terminal point becomes an equilibrium when a zero terminal tangent is imposed.
This property is local to the generated spline and should not be interpreted as a task-level stability or success guarantee.
The same spline output further supports uncertainty propagation and can be combined with classical controllers, such as null-space collision avoidance, without retraining the policy backbone.

The experiments support this output-interface view.
Low-dimensional studies isolate the flow-field mechanism and uncertainty propagation.
Simulated manipulation benchmarks evaluate the spline output under matched backbones, showing that the compact spline representation can keep task scores in a comparable range while reducing policy-output dimensionality and measured network-level forward FLOPs.
Real-robot case studies illustrate how the same spline object can be deployed with visual replanning, disturbance recovery, collision-avoidance control integration, and different policy backbones.
Together, these results suggest that SP provides a practical interface for connecting expressive learned policies with geometric and control-relevant motion structure.

SP also has limitations.
It changes the representation of the predicted action object, but it does not remove the need for an accurate and expressive policy backbone.
If the policy predicts an inappropriate spline, the structured decoder or the induced flow field cannot by itself guarantee task success.
The distance-to-spline corrective property applies to the generated motion under the assumptions of the analytical construction, not to arbitrary task objectives or arbitrary off-manifold states.
SP may also be less suitable for highly discontinuous or dynamic interactions, such as hitting a moving object, where additional task-specific modeling or engineering may be required.
Finally, although the trajectory-decoding view can use other spline families, the analytical flow-field realization in this work relies on concatenated quadratic curves with $C^0$ and $C^1$ continuity on the resulting spline (incl. junctions).
Extending the construction to broader spline families and evaluating uncertainty-aware or constraint-aware execution policies are important directions for future work.

\bibliography{refs}

\section*{Acknowledgment}
We used LLMs (e.g., ChatGPT) to help with paraphrasing and language polishing.
All technical content was written and verified by the authors.

\clearpage

\section{Appendix}

\subsection{From Spline to Distance Field}
\label{app:spline_to_distance_field}

We follow the notation in Sec.~\ref{sec:flow_field}. 
Given a query state $\mathbf{x}$, we compute its closest-point projection onto the decoded spline $\mathbf{f}_\theta(t)$, denoted by
\[
\bigl(t_\theta(\mathbf{x}),\mathbf{x}_{\mathrm{proj}}\bigr)
=
\mathcal{P}_{\mathbf{f}_\theta}(\mathbf{x}).
\]

Assume that $\mathbf{f}_\theta$ is a piecewise quadratic spline with segments $\mathbf{f}_{\theta,i}(\tau_i)$, $\tau_i\in[0,1]$, and quadratic Bernstein control points $\mathbf{w}_{i}^{1},\mathbf{w}_{i}^{2},\mathbf{w}_{i}^{3}$, all contained in $\mathbf{w}_\theta(\mathbf{o})$. 
For each segment $i$, the closest-point problem minimizes
\begin{equation}
    c_i(\tau_i) 
    =
    \frac{1}{2}
    \bigl(\mathbf{f}_{\theta,i}(\tau_i)-\mathbf{x}\bigr)^\top
    \bigl(\mathbf{f}_{\theta,i}(\tau_i)-\mathbf{x}\bigr)
\label{eq:app_segment_objective}
\end{equation}
with respect to $\tau_i$. 
At an interior stationary point, differentiating $c_i(\tau_i)$ and setting the derivative to zero gives
\begin{equation}
    \bigl(\mathbf{f}_{\theta,i}(\tau_i)-\mathbf{x}\bigr)^\top 
    \frac{d\mathbf{f}_{\theta,i}}{d\tau_i}(\tau_i)
    =
    0 .
\label{eq:app_projection_orthogonality}
\end{equation}
For the quadratic Bernstein segment,
\begin{equation}
\begin{aligned}
    \mathbf{f}_{\theta,i}(\tau_i)
    &=
    (1-\tau_i)^2 \mathbf{w}_{i}^{1} 
    + 2(1-\tau_i)\tau_i \mathbf{w}_{i}^{2} 
    + \tau_i^2 \mathbf{w}_{i}^{3}, \\
    \frac{d\mathbf{f}_{\theta,i}}{d\tau_i}(\tau_i)
    &=
    -2(1-\tau_i) \mathbf{w}_{i}^{1} 
    + (2-4\tau_i) \mathbf{w}_{i}^{2} 
    + 2\tau_i \mathbf{w}_{i}^{3}.
\end{aligned}
\label{eq:app_quadratic_bernstein}
\end{equation}
Substituting Eq.~\eqref{eq:app_quadratic_bernstein} into Eq.~\eqref{eq:app_projection_orthogonality} yields a cubic equation
\begin{equation}
    \alpha_i^{3} \tau_i^{3} 
    + \alpha_i^{2} \tau_i^{2} 
    + \alpha_i^{1} \tau_i 
    + \alpha_i^{0} 
    =
    0,
\label{eq:app_cubic_projection}
\end{equation}
where the coefficients $\alpha_i^{3},\alpha_i^{2},\alpha_i^{1},\alpha_i^{0}$ are determined by $\mathbf{w}_{i}^{1},\mathbf{w}_{i}^{2},\mathbf{w}_{i}^{3}$ and $\mathbf{x}$.

The candidate set for segment $i$ consists of all stationary roots in $[0,1]$ together with the two boundary points:
\begin{equation}
    \mathcal{C}_i(\mathbf{x})
    =
    \bigl\{
    \tau_i \in [0,1]
    \mid
    \tau_i \ \text{satisfies Eq.~\eqref{eq:app_cubic_projection}}
    \bigr\}
    \cup
    \{0,1\}.
\label{eq:app_candidate_set}
\end{equation}
The segment-wise closest distance is then
\begin{equation}
    d_{\theta,i}(\mathbf{x})
    =
    \min_{\tau_i^\ast \in \mathcal{C}_i(\mathbf{x})}
    \left\|
    \mathbf{f}_{\theta,i}(\tau_i^\ast)-\mathbf{x}
    \right\|.
\label{eq:app_segment_distance}
\end{equation}
For a piecewise spline with $K$ segments, the overall distance is
\begin{equation}
    d_\theta(\mathbf{x})
    =
    \min_{i=1,\ldots,K}
    d_{\theta,i}(\mathbf{x}).
\label{eq:app_overall_distance}
\end{equation}
The minimizer in \eqref{eq:app_overall_distance} defines the closest point $\mathbf{x}_{\mathrm{proj}}$ on the piecewise spline, so equivalently
$d_\theta(\mathbf{x})=\|\mathbf{x}-\mathbf{x}_{\mathrm{proj}}\|$.

When the closest-point projection is unique and differentiable at $\mathbf{x}$, and $d_\theta(\mathbf{x})>0$, the distance-field gradient is
\begin{equation}
    \nabla d_\theta(\mathbf{x})
    =
    \frac{
    \mathbf{x}-\mathbf{x}_{\mathrm{proj}}
    }{
    \left\|
    \mathbf{x}-\mathbf{x}_{\mathrm{proj}}
    \right\|
    }.
\label{eq:app_distance_gradient}
\end{equation}
We denote this unit vector by $\mathbf{n}_\theta(\mathbf{x})$, as in Sec.~\ref{sec:flow_field}.

\subsection{Flow Field Corrective Analysis}
\label{app:flow_field_stability}

We analyze the local corrective behavior of the flow field around the generated spline. 
The analysis applies at regular points where the closest-point projection is unique and differentiable, and where the active projection does not switch between spline segments. 
It characterizes distance-to-spline non-increase, rather than global task-level stability.

The induced dynamics are
\begin{equation}
    \dot{\mathbf{x}}
    =
    \mathbf{F}_\theta(\mathbf{x})
    =
    \alpha(\mathbf{x})\;
    \mathbf{n}_\theta(\mathbf{x})
    +
    \beta(\mathbf{x})\;
    \dot{\mathbf{f}}_\theta
    (\mathbf{x}).
\label{eq:app_flow_dynamics}
\end{equation} 
Consider the squared distance to the generated spline,
\begin{equation}
    V(\mathbf{x})
    =
    \frac{1}{2}d_\theta^2(\mathbf{x}).
\label{eq:app_distance_lyapunov}
\end{equation}
 Using 
$\dot d_\theta(\mathbf{x})
=
\mathbf{n}_\theta(\mathbf{x})^\top\dot{\mathbf{x}}$
at regular points with $d_\theta(\mathbf{x})>0$, we obtain
\begin{equation}
\begin{aligned}
    \dot V(\mathbf{x})
    &=
    d_\theta(\mathbf{x})\,\dot d_\theta(\mathbf{x})\\
    &=
    d_\theta(\mathbf{x})\,
    \mathbf{n}_\theta(\mathbf{x})^\top
    \dot{\mathbf{x}}\\
    &=
    d_\theta(\mathbf{x})\,
    \mathbf{n}_\theta(\mathbf{x})^\top
    \left(
    \alpha(\mathbf{x})\,\mathbf{n}_\theta(\mathbf{x})
    +
    \beta(\mathbf{x})\,
    \dot{\mathbf{f}}_\theta
    (\mathbf{x})
    \right)\\
    &=
    \alpha(\mathbf{x})\,d_\theta(\mathbf{x})
    +
    \beta(\mathbf{x})\,d_\theta(\mathbf{x})\,
    \mathbf{n}_\theta(\mathbf{x})^\top
    \dot{\mathbf{f}}_\theta
    (\mathbf{x}),
\end{aligned}
\label{eq:app_vdot}
\end{equation}
where we used $\|\mathbf{n}_\theta(\mathbf{x})\|=1$.

The first term in Eq.~\eqref{eq:app_vdot} is non-positive because $\alpha\le0$ and $d_\theta(\mathbf{x})\ge0$. 
We now consider the progression term.

\textbf{Interior projection:}
If the closest point lies in the interior of a differentiable spline segment, the closest-point condition gives
\begin{equation}
    \mathbf{n}_\theta(\mathbf{x})^\top
    \dot{\mathbf{f}}_\theta
    (\mathbf{x})
    =
    0.
\label{eq:app_interior_orthogonality}
\end{equation}
Thus, the progression term does not change the distance to the spline locally, and
$\dot V(\mathbf{x})=\alpha d_\theta(\mathbf{x})\le0$.

\textbf{Boundary projection:}
If the closest point lies at the terminal boundary $t_\theta(\mathbf{x})=T$, the defined weight $\beta(\mathbf{x})$
makes the progression term vanish. 
If the closest point lies at the start boundary $t_\theta(\mathbf{x})=0$, the one-sided closest-point condition gives
\begin{equation}
    \mathbf{n}_\theta(\mathbf{x})^\top
    \dot{\mathbf{f}}_\theta(0)
    \le
    0,
\label{eq:app_start_boundary_condition}
\end{equation}
so the progression term is also non-positive at this boundary.

Under these regularity and boundary assumptions, 
\begin{equation}
    \dot V(\mathbf{x}) \le 0.
\end{equation}
Therefore, the induced dynamics do not increase the distance to the generated spline. % at regular points. 

% When $\alpha(d)>0$ for $d>0$, the attraction term decreases the distance to the spline away from the curve. 
% The terminal point
% \begin{equation}
%     \mathbf{x}^{\star}
%     =
%     \mathbf{f}_\theta(T)
% \end{equation}
% is an equilibrium of the constructed field when the zero-terminal-tangent condition is imposed:
% \begin{equation}
%     \dot{\mathbf{f}}_\theta(T)=\mathbf{0},
%     \qquad
%     \mathbf{F}_\theta(\mathbf{x}^{\star})=\mathbf{0}.
% \end{equation}

\subsection{Real-World Setup and Backbone Deployment Details}
\label{sec:realworld_details}
\label{sec:realworld_backbone_details}

\begin{figure}[h]
    \centering
    \includegraphics[width=1.0\linewidth]{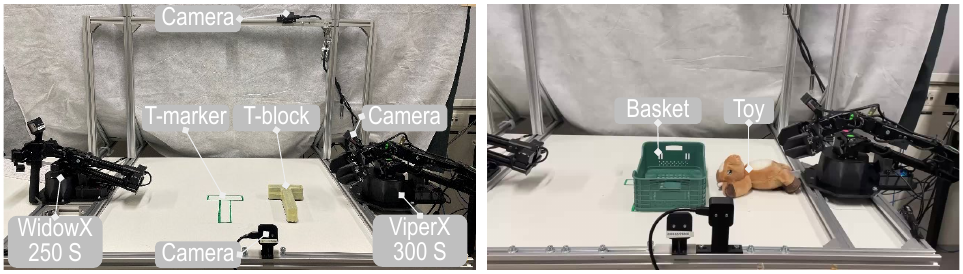}
    \caption{
    ALOHA teleoperation platform for the PushT and toy packing tasks.
    }
    \label{fig:aloha_teleoperation}
\end{figure}

\textbf{Robot platform and data collection:}
The real-world experiments are conducted on an ALOHA-style bimanual teleoperation platform~\cite{zhao2023learning}, configured with a ViperX300s follower arm and a WidowX250 leader arm.
The setup also includes three RealSense D405 cameras providing top-down, front, and wrist views, as shown in Fig.~\ref{fig:aloha_teleoperation}.
During data collection, the operator controls the follower arm by teleoperating the leader arm.
The cameras record RGB images at $640 \times 480$ resolution and 30 Hz.
The dataset is stored in the LeRobotDatasetV3.0 format and contains robot actions, motor states, RGB observations, and language embeddings.
We collect demonstrations for two real-world manipulation tasks: PushT and toy packing.
In PushT, the robot pushes a T-shaped block to a target configuration.
We collect 100 demonstrations with randomized initial object poses, resulting in approximately 20,000 training samples after filtering.
In toy packing, the robot grasps a deformable toy sheep from the table and places it into a basket.
We collect 69 demonstrations, resulting in over 37,000 RGB image-action pairs before filtering; the filtered training set contains approximately 1,000 samples.
Each demonstration contains approximately 400--700 timesteps.

\textbf{Backbones and training details:}
We instantiate Spline Policy (Traj.) with multiple real-world policy backbones, including ACT, DP, and the VLA model PI05.
The resulting policies are denoted SP-ACT, SP-Diff, and SP-PI05.
In all cases, SP changes the trajectory-output representation to spline parameters while preserving the corresponding policy backbone.
ACT and Diffusion are trained from scratch, while PI05 is fine-tuned from a pretrained base model.
All policies are trained using a single A100 GPU on the Alvis cluster.
The training configurations are summarized in Table \ref{training_parameters}.

\begin{table}[h]
    \centering
    \caption{Training parameters for real-world backbone deployment.}
    \label{training_parameters}
    \begin{tabular}{@{} >{\raggedright\arraybackslash}p{0.15\textwidth}
        >{\raggedright\arraybackslash}p{0.09\textwidth}
        >{\raggedright\arraybackslash}p{0.09\textwidth}
        >{\raggedright\arraybackslash}p{0.09\textwidth}
        @{} }     
        \toprule
        \textbf{Parameters}   
        & \textbf{SP-ACT} 
        & \textbf{SP-Diff} 
        & \textbf{SP-PI05} \\
        \midrule
        Step (K) & 100 & 100 & 3 \\
        Optimizer & AdamW & Adam & AdamW \\
        Learning rate ($\times 10^{-5}$) & 1 & 10 & 2.5 \\
        Batch size & 8 & 8 & 32 \\
        Loss function & L1 & L2 & L2 \\
        Time (h) & 2.1 & 1.5 & 3.9 \\
        Model size & 52M & 270M & 4B \\
        \bottomrule
    \end{tabular} 
\end{table}

\textbf{Evaluation protocol and results:}
Each policy is evaluated over 10 trials with randomized object poses.
The success rates are reported in Table \ref{tab:realworld_success_rate}.
SP-PI05 achieves 10/10 success on PushT and 9/10 on toy packing.
SP-Diff achieves 6/10 and 8/10 success on PushT and toy packing, respectively.
SP-ACT is less reliable in these trials, especially on PushT, where lagging and stuck behaviors are occasionally observed.
Because the compared backbones differ in model size, pretraining, optimizer, and training schedule, these results should be interpreted as deployment examples across different backbones rather than as a controlled comparison of backbone architectures.

\begin{table}[h]
\centering
\caption{Real-world success rates on PushT and toy packing.}
\label{tab:realworld_success_rate}
\begin{tabular*}{\columnwidth}{@{\extracolsep{\fill}}lccc}
\toprule
 & \textbf{SP-ACT} & \textbf{SP-Diff} & \textbf{SP-PI05} \\
\midrule
PushT & 0/10 & 6/10 & 10/10 \\
Toy packing & 5/10 & 8/10 & 9/10 \\
\bottomrule
\end{tabular*}

\vspace{2pt}
{\footnotesize
All methods use Spline Policy (Traj.) with different policy backbones.
Each entry reports the number of successful trials out of 10.
}
\end{table}

\end{document}